\documentclass{article}
\usepackage[english]{babel}
\usepackage[letterpaper,top=2cm,bottom=2cm,left=3cm,right=3cm,marginparwidth=1.75cm]{geometry}
\usepackage{amsmath}

\usepackage{graphicx}
\usepackage[colorlinks=true, allcolors=black]{hyperref}
\usepackage{booktabs}
\usepackage{siunitx}
\usepackage{algorithm}
\usepackage{algpseudocode}
\usepackage{float} 
\usepackage{bm}
\usepackage{enumitem}

\title{\textsc{$\Phi$asmaNet}: A Staged Spectral Token-Refinement Architecture for Fast\\
Gravitational-Wave Signal-to-Noise Ratio Prediction}

\author{\normalsize Ioannis Tsioulis$^{1,2}$, Anastasios Tefas$^{3}$,  Nikolaos Stergioulas$^{1}$\\
\small ${}^1$Department of Physics, Aristotle University of Thessaloniki, Greece \\
\small ${}^2$School of Computing, Mediterranean College (University of Derby), Athens, Greece \\
\small ${}^3$Department of Informatics, Aristotle University of Thessaloniki, Greece
}

\usepackage{amssymb}
\usepackage{rotating}

\date{October 31, 2026}

\begin{document}
\maketitle

\begin{abstract}
Quantifying selection effects in gravitational-wave population studies requires the
signal-to-noise ratio (SNR) of millions of simulated compact binaries, each conventionally
obtained by generating a waveform and computing a noise-weighted inner product, a cost
that is modest per source but becomes prohibitive in aggregate at the scale of
$10^{6}$--$10^{7}$ injections. We present
\textsc{$\Phi$asmaNet} (from Greek \emph{$\varphi\acute{\alpha}\sigma\mu\alpha$}, spectrum), a
compact neural architecture for structured tabular regression whose novelty is to make
\emph{staged spectral} representation refinement an explicit architectural prior: learned input
tokens are refined by an ordered cascade of attention-based \emph{$\Phi$-Stages}, each
augmented by the new \emph{$\Phi$-Refractor} subspace-mixing module, and then revisited
through gated top-down \emph{$\Phi$-Feedback} iterations. Instantiated as a surrogate mapping the $11$-dimensional
parameters of precessing, higher-mode binary black hole signals (IMRPhenomXPHM) to the
phase-averaged optimal SNR in a single detector at fixed Advanced LIGO design sensitivity,
\textsc{$\Phi$asmaNet} outperforms MLP, 1D-ResNet, and Transformer-encoder baselines under a
matched training protocol at every dataset scale from $10^{5}$ to $10^{6}$ samples, while
using only ${\sim}\,732$k parameters. At $10^{6}$ samples it reaches
$100-\mathrm{MAPE}=98.73\%$ on the held-out test set, $99.29\%$ after a rank-preserving
monotonic (PCHIP) calibration, and $99.39\%$ ($R^{2}=0.9998$) in the detection-relevant
regime $\rho>8$, with accuracy stable across independently generated dataset realizations.
Sustaining ${\sim}\,1.2\times10^{5}$ SNR evaluations per second on a single GPU, which processes a
$10^{6}$-injection catalogue end-to-end in ${\sim}\,8.3$~s, about two orders of magnitude faster than
the conventional waveform-and-inner-product calculation on a 16-core CPU, and exactly
rescalable in distance through $\rho\propto1/d_L$, the surrogate is fast enough for the
repeated, catalogue-wide detectability evaluations that selection-effect studies demand.
The fully documented pipeline is intended as a reproducible benchmark for future SNR
surrogates, and
out-of-the-box results on public tabular benchmarks indicate that the staged inductive
bias remains competitive beyond the gravitational-wave setting. Although demonstrated here for a
fixed single-detector configuration, the construction extends directly to detector
networks and epoch-specific sensitivities, positioning staged neural surrogates as a
practical route to keeping selection-effect estimation tractable as detector upgrades
raise event rates, and with them injection-campaign demands, by orders of magnitude.
\end{abstract}

\section{Introduction}
The field of Gravitational-Wave (GW) astronomy, ushered in by the Advanced LIGO,
Advanced Virgo, and KAGRA detectors (collectively the LVK Collaboration)
\cite{aasi2015advancedligo,acernese2014advanced,akutsu2020overview}, has opened an
unprecedented observational window onto the dynamics of compact objects, principally
through the detection of Compact Binary Coalescences (CBCs), beginning
with GW150914 \cite{abbott2016observation}. As successive observing runs have enlarged the
catalogue of confirmed events from a handful of detections to of order four hundred across
the Gravitational-Wave Transient Catalogs
\cite{abbott2019gwtc1,abbott2021gwtc2,abbott2024gwtc21,abbott2023gwtc3,lvk2025gwtc4,lvk2026gwtc5},
the scientific emphasis has shifted from the
detection and characterization of \emph{individual} sources towards the statistical
inference of the underlying astrophysical \emph{population} of mergers. This shift places a
new computational demand at the heart of the analysis: population properties can only be
recovered without bias once the survey's selection effects, the fact that louder,
nearer, and more favourably oriented sources are preferentially detected, are accurately
quantified and corrected for \cite{gerosa2020gravitational,mould2022deep}. Because the next
generation of detector upgrades is expected to raise the event rate by orders of magnitude,
the methods used to evaluate these selection effects must remain efficient when scaled to
very large synthetic populations.

Quantifying selection effects reduces, at its core, to repeatedly asking whether a source
with given parameters would have been detected, a question that is almost universally
answered through its signal-to-noise ratio (SNR). Estimating the detectable fraction of a
proposed population, and doing so afresh at every likelihood evaluation of a hierarchical
Bayesian analysis, therefore requires the SNR (or a detectability proxy derived from it) to
be evaluated for millions of synthetic sources
\cite{gerosa2020gravitational,chapman2023rapid,callister2024}. As each such evaluation
conventionally entails generating a waveform and computing a noise-weighted inner product,
the cumulative cost can grow prohibitive at the scale of a full population analysis. This is
the gap that fast, accurate surrogate models for the SNR are designed to close, and it is
the setting in which the present work is positioned.

Building on preliminary efforts to accelerate SNR computations for binary
black hole (BBH) injections using simulated datasets \cite{tsioulis2023rapid}, this work introduces
\textsc{$\Phi$asmaNet}, a compact neural surrogate that enforces staged spectral representation
refinement through a cascade-based architectural organization. The SNR is computed as phase-averaged
optimal for a single detector assuming a specific waveform approximant and a fixed representative
noise power spectral density (PSD). Each waveform, for which the SNR is calculated, depends on 11
parameters, when projected onto the antenna pattern of the chosen detector. Our model is intended for
high-throughput \emph{simulation-driven} applications under matched assumptions, and not as a drop-in
replacement for end-to-end low-latency search pipelines.

Evaluated on datasets from 100k to 1M BBH simulations, \textsc{$\Phi$asmaNet} achieves the best
performance \emph{among several different evaluated neural architectures}. At 1M samples, it reaches
an accuracy (defined as $100-\mathrm{MAPE}$) of $98.73\%$, rising to $99.29\%$
after a rank-preserving deploy-time calibration, with the post-calibration accuracy improving
monotonically with sample size, while sustaining ${\sim}\,1.2\times10^{5}$ SNR evaluations per second on a single
GPU, which processes a $10^{6}$-injection catalogue end-to-end in ${\sim}\,8.3$~s, about two orders of
magnitude faster than the conventional waveform-and-inner-product calculation on a 16-core CPU.
Creating different realizations of the training and test datasets, we find very consistent
accuracy results. Here, we focus on the BBH SNR regression setting under specific assumptions.
Broader generalization (e.g., varying PSDs and detector networks) is left for future work.

While existing architectures can learn hierarchical representations implicitly through depth or
attention mechanisms, \textsc{$\Phi$asmaNet}
encodes staged representation refinement as an explicit architectural prior,
organizing standard neural components into non-exchangeable $\Phi$-Stages. Our results are \emph{consistent with} the view that the staged
organization of \textsc{$\Phi$asmaNet} can act as a useful inductive bias for this surrogate task under the
matched protocol used here. However, more exhaustive ablations would be required to attribute gains
to individual components.

Our paper is organized as follows. Section~\ref{Sec: challenge} reviews the computational bottlenecks in direct SNR evaluation and motivates fast surrogates. Section~\ref{sec:related} summarizes recent ML/DL approaches for accelerating SNR and related GW inference tasks. Section~\ref{sec:methodology} describes our dataset construction and experimental protocol. Section~\ref{sec:classic} introduces the baseline deep learning architectures used for comparison, while Section~\ref{sec:phasmanet} presents \textsc{$\Phi$asmaNet}. Section~\ref{sec:novelty} discusses the architectural design choices and their motivation. Section~\ref{sec:training} details the training protocol and hyperparameter selection. Section~\ref{sec:architecture_comparison} reports the scaling results across dataset sizes and compares model performance under a matched evaluation procedure. Section~\ref{sec:best-model-diagnostics} presents diagnostics for the best model: training stability, residual structure, accuracy in the detection-relevant regime, and end-to-end inference timing, together with the mapping from SNR predictions to detection probability (Section~\ref{sec:popdet}). Section~\ref{sec:discussion} summarizes our findings and outlines future directions. Finally, Appendix~\ref{sec:appendix} reports out-of-the-box results on standard public tabular benchmarks, probing the transferability of the architecture beyond the gravitational-wave setting.

\begin{figure}[t]
  \centering
  \makebox[\textwidth][c]{%
    \includegraphics[
      width=0.9\textwidth,
      height=0.5\textheight,
      keepaspectratio
    ]{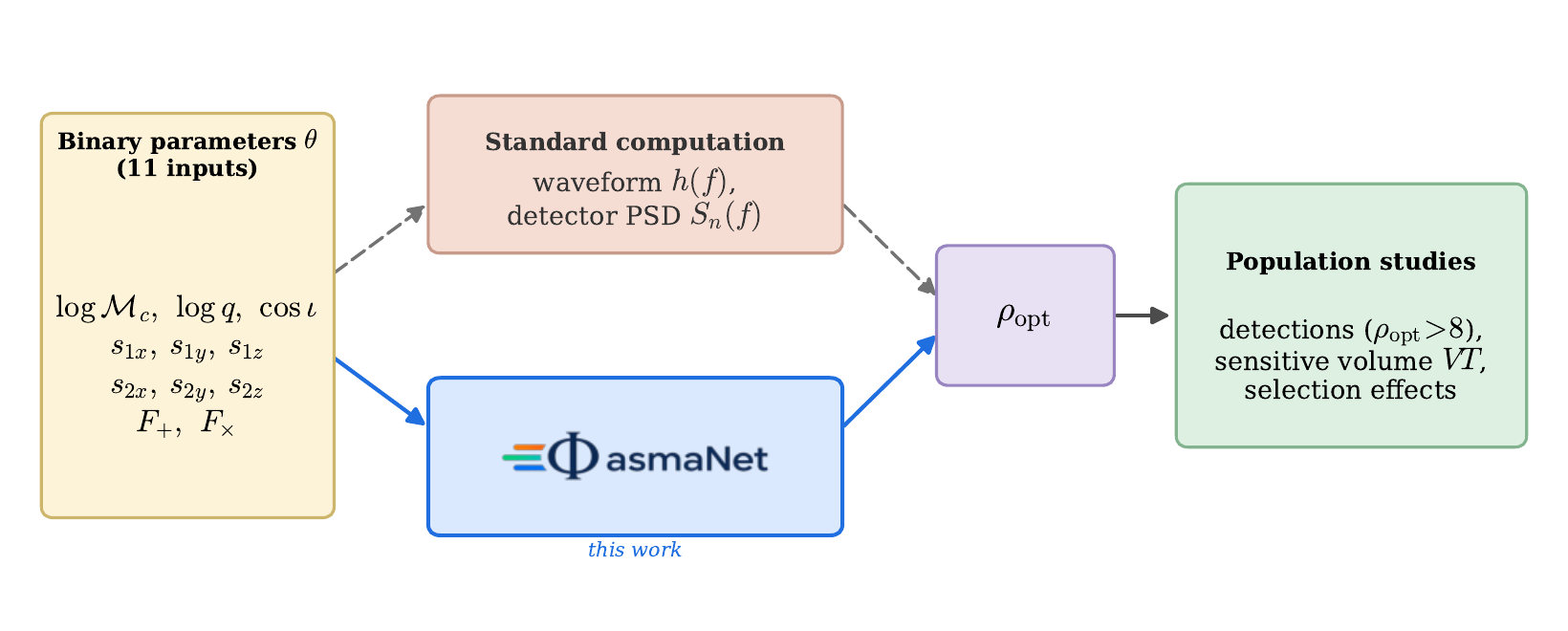}%
  }
\caption{Role of the SNR surrogate. For a set of binary parameters $\theta$, the optimal
 SNR $\rho_{\rm opt}$ governs detectability. Evaluating it directly for a large number of injections is computationally demanding and can become a bottleneck in population and
selection-effect studies that require very large numbers of evaluations.
\textsc{$\Phi$asmaNet} replaces this per-source computation with a neural-network surrogate
that maps $\theta$ directly to $\rho_{\rm opt}$, enabling high-throughput injection
campaigns in which the detectability ($\rho_{\rm opt}>8$), the sensitive volume
$V\!T$, and selection effects are evaluated for large populations.}

\label{fig:schematic}
\end{figure}

\section{The Challenge of Signal-to-Noise Ratio Calculation}
\label{Sec: challenge}
A foundational requirement for almost all GW data analysis tasks including detection significance
estimation, performing accurate parameter estimation, and executing large-scale population
studies is the frequent and accurate calculation of the Signal-to-Noise Ratio (SNR),
$\rho(\bm{\theta})$. The SNR quantifies the strength of a potential GW signal, characterized by a
vector of parameters $\bm{\theta}$, against the detector's background noise PSD $S_n(f)$. This
calculation involves a noise-weighted integral of the waveform $\tilde{h}(f;\bm{\theta})$ in the
Fourier domain:
\begin{equation}
    \rho(\bm{\theta}) = \left[ 4 \int_{f_{\min}}^{f_{\max}}
    \frac{\tilde{h}^\ast(f;\bm{\theta})\,\tilde{h}(f;\bm{\theta})}{S_n(f)} \, df \right]^{1/2}.
    \label{eq:snr}
\end{equation}

\begin{figure}[t]
  \centering
  \makebox[\textwidth][c]{%
    \includegraphics[
      width=0.9\textwidth,
      height=0.5\textheight,
      keepaspectratio
    ]{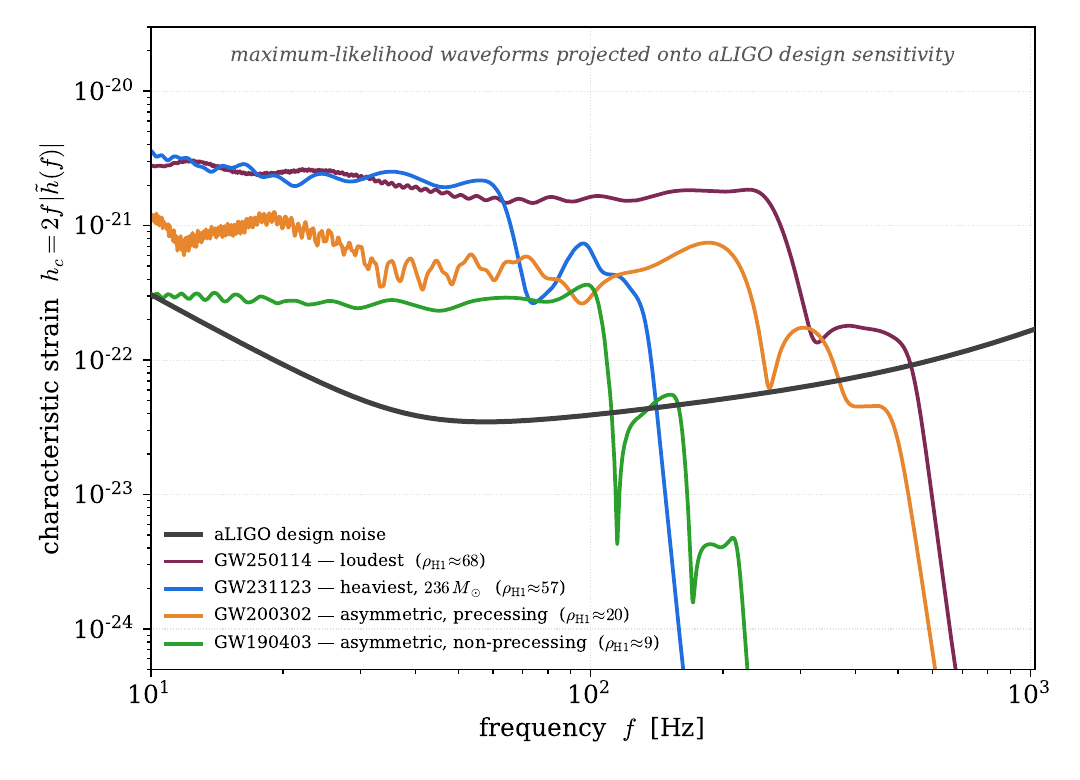}%
  }
\caption{Four real GWTC events reconstructed with the same waveform model and noise
curve used to build the training set. Each curve is the characteristic strain
$h_c=2f|\tilde h(f)|$ of the maximum-likelihood posterior sample
(IMRPhenomXPHM, projected onto H1 at the event's sky position and GPS time). The
black curve is the aLIGO \emph{design} noise $h_n=\sqrt{f S_n(f)}$.  The events span the loudest,
heaviest, and two unequal-mass systems with and without precession.
}

\label{fig:real_events}
\end{figure}

To illustrate the regime the surrogate operates in, Fig.~\ref{fig:real_events}
shows four real events from the Gravitational-Wave Transient Catalogs~\cite{lvk2025gwtc4,lvk2026gwtc5}, chosen to span its extremes: the
loudest detection to date (GW250114~\cite{lvk2025gw250114}), the heaviest binary
(GW231123, $236\,M_\odot$~\cite{lvk2025gw231123}), and two strongly unequal-mass systems that contrast a
precessing configuration (GW200302, $\chi_p\!\approx\!0.5$~\cite{abbott2023gwtc3}) with an
aligned-spin one (GW190403, $\chi_{\rm eff}\!\approx\!0.7$, $\chi_p\!\approx\!0.1$~\cite{abbott2024gwtc21}).
For each event we take the maximum-likelihood sample of the public posterior data
release and reconstruct the frequency-domain signal with exactly the configuration
used to generate the training set (IMRPhenomXPHM, $f_{\rm low}=10$~Hz,
$f_{\rm Nyq}=1024$~Hz, $\Delta f=1/20$~Hz), projecting onto the LIGO-Hanford
detector at the event's inferred sky position and coalescence time. Plotted as
characteristic strain against the aLIGO design noise curve, the four signals make
the meaning of the SNR concrete: the area by which each curve rises above the
noise is its $\rho^2$, ranging here from $\rho_{\rm H1}\!\approx\!68$ for the
loudest system down to $\rho_{\rm H1}\!\approx\!9$ for the most distant. The
visible amplitude modulations are genuine higher-order-mode and spin-precession
features, precisely the structure, absent from non-precessing dominant-mode
models, that the IMRPhenomXPHM-based surrogate must reproduce across the parameter
space. We emphasise that these waveforms are projected onto a single detector at
fixed design sensitivity rather than the actual detector network and noise, so
$\rho_{\rm H1}$ differs from the network SNRs at which the events were observed.

While the direct evaluation of this integral for a single event is feasible, the necessary waveform
construction can be time-consuming. In particular, high-fidelity waveform models, such as
effective-one-body (EOB) models, often necessitate the integration of differential equations.
Even the closed-form, frequency-domain phenomenological model used to generate
our labels (IMRPhenomXPHM) is not free of this cost: capturing generic spin precession and
higher-order modes requires assembling and rotating many harmonic contributions, each evaluated on
the fine frequency grid ($\Delta f = 1/20$~Hz up to $1024$~Hz, i.e.\ ${\sim}2\times10^{4}$ bins)
needed to resolve the in-band signal, and this per-source cost is incurred afresh at every one of the
millions of evaluations a population analysis requires. This
computational cost is felt most acutely when detectability must be established
for an entire synthetic population. Robust population models require large synthetic catalogs,
often millions of sources, to quantify selection effects and detection efficiency
\cite{mould2022deep,mould2023one}. In a hierarchical Bayesian analysis the detectable fraction of
such a catalog must moreover be recomputed at every likelihood evaluation, so that even individually
inexpensive SNR calculations accumulate into a prohibitive total. This motivates fast surrogate
models tailored to the same fixed analysis configuration used to generate the labels.
Furthermore, conventional computational methods, such as simple polynomial interpolation or look-up
tables, become inefficient or inappropriate once the intrinsic parameter space includes generic, precessing spins, for which the SNR no longer factorizes into a simple intrinsic normalization times an analytic distance–orientation scaling \cite{gerosa2020gravitational,wong2020gravitational}.

Deep Learning (DL) presents a robust alternative to overcome these computational hurdles. Feed-forward
Neural Networks (NNs), specifically Multilayer Perceptrons (MLPs), are mathematically established as
universal approximators capable of accurately modeling virtually any continuous, complex function over
a defined parameter space. Beyond standard MLPs, more advanced architectures such as Convolutional
Neural Networks (CNNs), residual networks (ResNets), and Transformer models can also be employed,
especially when the SNR function exhibits structured dependencies (e.g., local correlations,
multi-scale behavior, or attention-like relationships among parameters).

The core strategy is to replace the slow, direct calculation of the SNR function, $\rho(\bm\theta)$,
with a fast, accurate interpolation performed by a pre-trained NN. Although the initial training phase
can be computationally intensive, once the chosen architecture (MLP, CNN, ResNet, or Transformer) is
trained and validated for sufficient accuracy, the computational cost associated with a single
prediction (inference) is constant and significantly lower than solving the complex integral required
by Eq.~(\ref{eq:snr}). This approach effectively transforms the challenging problem of numerically
evaluating an integral in a high-dimensional space into an efficient function evaluation, while
allowing the use of more expressive and specialized network architectures when beneficial.

\section{Related Work}
\label{sec:related}
Deep learning has by now been applied across essentially the entire gravitational-wave
analysis chain: signal detection, waveform modelling, parameter estimation, and
population inference \cite{mould2022deep,gerosa2020gravitational}. The strand most relevant
to the present work is the use of neural networks as fast \emph{interpolants}: surrogates
that replace an expensive physical computation with a cheap, pre-trained function
evaluation. We organise the prior art into three overlapping themes and then position
\textsc{$\Phi$asmaNet} against it.

\paragraph{Fast surrogates for the SNR.}
The proposal to replace the noise-weighted inner product of Eq.~(\ref{eq:snr}) with a
learned interpolant was introduced by Wong, Ng and Berti \cite{wong2020gravitational}, who
trained a simple multilayer perceptron to map the four intrinsic parameters of
aligned-spin binaries $(M_{\rm tot}, q, \chi_1, \chi_2)$ to the optimal SNR. Their network
evaluated $10^6$ SNRs in roughly one minute on a four-core CPU, with median fractional
errors below $10^{-3}$ and speed-ups of $120$--$7.5\times10^{4}$ relative to direct waveform
generation, while compressing a $\sim\!100$~MB look-up table into a $\sim\!5$~MB model.
These figures illustrate why SNR interpolation, although conceptually simple, is an
attractive target for DL. More recently, the \textsc{gwsnr} package \cite{gwsnr} has turned
this idea into a production tool, combining a partial-scaling interpolation of the optimal
SNR with an artificial-neural-network estimator and a hybrid recalculation scheme that falls
back to the exact inner product for sources near the detection threshold, where
approximation errors matter most.

\paragraph{Learned detectability and selection functions.}
A complementary line of work models the detection probability directly rather than the SNR.
Gerosa, Pratten and Vecchio \cite{gerosa2020gravitational} trained neural-network
\emph{classifiers} to predict whether a compact-binary signal would be detected, including
the effects of spin precession, higher-order modes, and multiple detectors that are awkward
to capture with a single semi-analytic SNR threshold. Mould, Gerosa and Taylor
\cite{mould2022deep} embedded such learned ingredients in a hierarchical Bayesian framework
for inferring black-hole populations. Most recently, Callister, Essick and Holz
\cite{callister2024} trained a neural-network emulator of the Advanced LIGO/Virgo selection
function on real pipeline injections, learning a continuous
$P(\mathrm{det}\,|\,\bm{\theta})$ across the $13$-dimensional space of binary masses, spins,
and extrinsic parameters, and demonstrating that it captures detection-pipeline effects that a fixed SNR threshold misses, such as
noise non-stationarity and signal-consistency cuts.
Such emulators are valuable precisely because detectability must be re-evaluated for a fresh
population at every step of a population-inference sampler, a setting in which a fast
surrogate removes an otherwise prohibitive cost. The same motivation drives space-based
applications: for Extreme Mass Ratio Inspirals (EMRIs) observable by LISA, Chapman-Bird,
Berry and Woan \cite{chapman2023rapid} chained an SNR-interpolating network with a second
network that learns the selection function, accelerating detectability estimation by up to
six orders of magnitude relative to direct computation. Underpinning all such
detectability models, the simple optimal-SNR threshold they refine has itself been calibrated against
observed catalogues: Mould, Moore and Gerosa \cite{mould2024calibrating} infer single-detector
thresholds that justify the $\rho>8$-type criteria routinely used to define detectable populations
\cite{mould2023one}.

\paragraph{Waveform-level surrogates.}
A distinct but related strategy interpolates the waveform itself, yielding the SNR as a
by-product. Magee et al.\ \cite{magee2024neural} trained networks to interpolate the
singular-value-decomposition reconstruction coefficients used by matched-filter search
pipelines, generating waveforms, and equivalently SNR time series, in milliseconds
and enabling low-latency enrichment of candidate events. In a related
data-driven variant, Ma et al.~\cite{ma2024using} replace the matched-filtering template bank
with a deep-learning model bank that predicts the waveform template embedded in the detector
strain, then recovers the matched-filtering SNR as a by-product by correlating that predicted
template with the data. Whereas these models reconstruct
the signal across a local patch of the template bank or directly from the
strain, our surrogate predicts the scalar
optimal SNR directly across the full projected parameter space from the
binary parameters alone.
\paragraph{This work.}
Against this backdrop, two observations motivate \textsc{$\Phi$asmaNet}. First, the SNR surrogates
above almost exclusively employ plain MLPs over low-dimensional intrinsic-parameter inputs.
The more expressive architectures that have reshaped structured/tabular regression,
residual connections and attention, remain largely unexplored for this
task. This gap persists even in the production package \textsc{gwsnr}
\cite{gwsnr}, the closest existing tool: its neural SNR estimator, trained on the same
IMRPhenomXPHM waveform family we adopt, is a compact perceptron with two $32$-unit hidden layers
that predicts the optimal SNR from only five engineered scalar inputs, in which the spin degrees of
freedom are compressed into a single effective spin $\chi_{\rm eff}$. \textsc{$\Phi$asmaNet} instead
regresses the SNR from the full $11$-dimensional parameter vector, retaining all six spin components,
through a deep staged-refinement network, allowing it to represent the genuine precession- and
higher-mode-induced structure that a $\chi_{\rm eff}$ parametrization cannot capture. Second, the quantity we target is the phase-averaged, single-detector optimal SNR for
a fixed waveform model and noise PSD, but evaluated over the full \emph{11-dimensional}
projected parameter vector, a substantially harder regression than the four-dimensional
aligned-spin setting of \cite{wong2020gravitational}. Crucially, our target is
the continuous optimal SNR itself rather than a particular pipeline's detection probability
$P(\mathrm{det}\,|\,\bm\theta)$ \cite{callister2024}: because the SNR is a deterministic physical
quantity, a single trained surrogate supports any downstream detection threshold, an exact distance
rescaling ($\rho\propto1/d_L$), and any noise-averaged $P_{\rm det}$ model, whereas a learned
$P_{\rm det}$ fixes these choices at training time. We characterise  \textsc{$\Phi$asmaNet}'s  accuracy and scaling
behaviour under a controlled, fixed-configuration protocol against representative neural
baselines (MLP, ResNet, and Transformer). We do not claim an exhaustive benchmark against
all strong tabular regressors. Our aim is a controlled, like-for-like comparison among
neural surrogates for this regression problem.

\section{Methodology}
\label{sec:methodology}

To develop an effective NN interpolator for the SNR $\rho(\bm{\theta})$ as a function of binary
parameters $\bm{\theta}$, we first construct a comprehensive training dataset that spans the relevant
parameter space of spinning BBH mergers (including generic spin orientations). This dataset is
generated using the PyCBC library \cite{biwer2019pycbc}, which enables efficient computation of
frequency-domain waveforms and optimal SNR values. The process leverages uniform
random sampling in the component masses (with a quasi-random Sobol scheme available as an alternative,
see below) and uniform random sampling for spins and extrinsic parameters, ensuring broad
and uniform coverage while respecting physical constraints. Below, we detail the parameter selection,
waveform generation, SNR computation, and dataset assembly.

\subsection{Parameter Space and Sampling Strategy}

The parameter vector $\bm \theta$ includes intrinsic binary properties: detector-frame chirp mass
$\mathcal{M}_c$, mass ratio $q \geq 1$, and dimensionless component spins $\boldsymbol{s}_{1,2}$,
alongside extrinsic orientation parameters such as the cosine of inclination $\cos\iota$, right
ascension $\alpha$, declination $\delta$, polarization angle $\psi$, coalescence time $t_c$, and
luminosity distance $d_L$. The masses are detector-frame (redshifted) quantities:
the waveform generator applies no cosmology, so the value passed sets the observed frequencies
directly, and a source-frame population at redshift $z$ must be queried with
$m^{\rm det}=(1+z)\,m^{\rm src}$. For this proof-of-concept, we focus on the LIGO-Hanford (H1)
detector with the Advanced LIGO zero-detuned high-power (aLIGOZeroDetHighPower) PSD. The full
parameter vector $\boldsymbol{\theta}$ includes both intrinsic parameters and extrinsic projections
(via the antenna pattern functions $F_{+}$ and $F_{\times}$), as well as distance and a
coalescence-time offset as tabulated features. In this configuration the luminosity
distance and coalescence time are held fixed at $d_L=1000$~Mpc and $t_c=0$. They are therefore
constant (zero-variance) quantities excluded from the model input rather than treated as informative features. Because the
optimal SNR scales exactly as $\rho\propto 1/d_L$, the labels computed at the reference distance can
be rescaled to any $d_L$ at query time (see Sec.~\ref{sec:popdet}), so fixing $d_L$ during generation
incurs no loss of generality. The low-frequency cutoff is $f_\mathrm{low} = 10$ Hz, with a sampling
rate of 2048 Hz and signal duration of 20 s, yielding a frequency resolution $\Delta f = 1/20$ Hz.

Mass parameters are drawn uniformly in the component masses,
$m_1,m_2 \sim \mathcal{U}(3,170)\,M_\odot$, with the pair reordered to enforce
$3 \le m_2 \le m_1 \le 170\,M_\odot$ (hence $q=m_1/m_2 \ge 1$). The chirp mass and mass ratio used as
features are then derived from each pair, spanning $\log_{10}q \in [0, \log_{10}(170/3)]$ and the
corresponding $\log_{10}\mathcal{M}_c$ range. This uniform-$(m_1,m_2)$ scheme requires no rejection
step, since every reordered draw satisfies the mass bounds by construction. As an alternative, a
quasi-random Sobol sampler \cite{sobol1967distribution} (via SciPy's QMC module) in
the log-uniform $(\log_{10}\mathcal{M}_c, \log_{10}q)$ plane is also provided. The desired number of valid draws $N_s$ ranges from $10^5$ to
$10^6$ samples in the experiments reported here. We verified that, for detection-oriented accuracy
($\rho>8$), the two samplers perform comparably across the mass range at $10^6$ samples, with the default
uniform-$(m_1,m_2)$ scheme being the more distribution-robust of the two. Spin magnitudes follow uniform
distributions in $[0, 0.99]$, with polar and azimuthal angles in $[0, \pi]$ and $[0, 2\pi]$,
respectively, converted to Cartesian components via spherical-to-Cartesian transformation.
We note that uniform sampling of the polar (tilt) angle over $[0,\pi]$ is not the
isotropic measure (which is uniform in $\cos\theta$). It mildly over-weights aligned configurations
and under-covers in-plane (precessing) spins. This is a deliberate training-coverage choice rather
than an astrophysical prior: the surrogate is a function $\rho(\bm\theta)$ evaluated on injections
drawn from the desired population, so the astrophysical spin distribution is imposed at query time
and the training measure only controls where the emulator is accurate. At $10^6$ samples the in-plane
region remains densely populated, so this choice costs the surrogate nothing. Extrinsic parameters
are uniformly sampled: $\cos\iota \in [-1,1]$, $\sin\delta \in [-1,1]$, $\alpha \in [0, 2\pi]$,
$\psi \in [0, \pi]$.

This yields 11 informative input features per sample ($\log M_c$, $\log q$,
$\cos\iota$, six spin components, and antenna patterns $F_{+}$, $F_{\times}$). The luminosity distance
and coalescence time are held fixed during data generation ($d_L=1000$~Mpc, $t_c=0$) and excluded from
the model input: $d_L$ is reintroduced analytically through the exact $\rho\propto1/d_L$ rescaling
(Sec.~\ref{sec:popdet}), and $t_c$ leaves the optimal SNR invariant. The sky position
and polarization enter only through the precomputed $F_{+},F_{\times}$ rather than
as raw angles. For reproducibility, the random-number generators are seeded with 42
(a NumPy \texttt{default\_rng(42)} stream for the masses and a \texttt{seed(42)} stream for the
remaining parameters).
The sampling ranges and priors used for each parameter are summarized in Table~\ref{tab:sampling}.

For precessing, higher-mode waveforms the coalescence phase $\phi_c$ induces a real,
structured variation in the SNR. For astrophysically realistic (LVK-inferred, low-spin) populations,
however, the resulting sensitive-volume bias is sub-percent.  As described
in Sec.~\ref{sec:waveform}, we marginalize $\phi_c$.

\begin{figure}[t]
  \centering
  \makebox[\textwidth][c]{%
    \includegraphics[
      width=0.8\textwidth,
      height=0.5\textheight,
      keepaspectratio
    ]{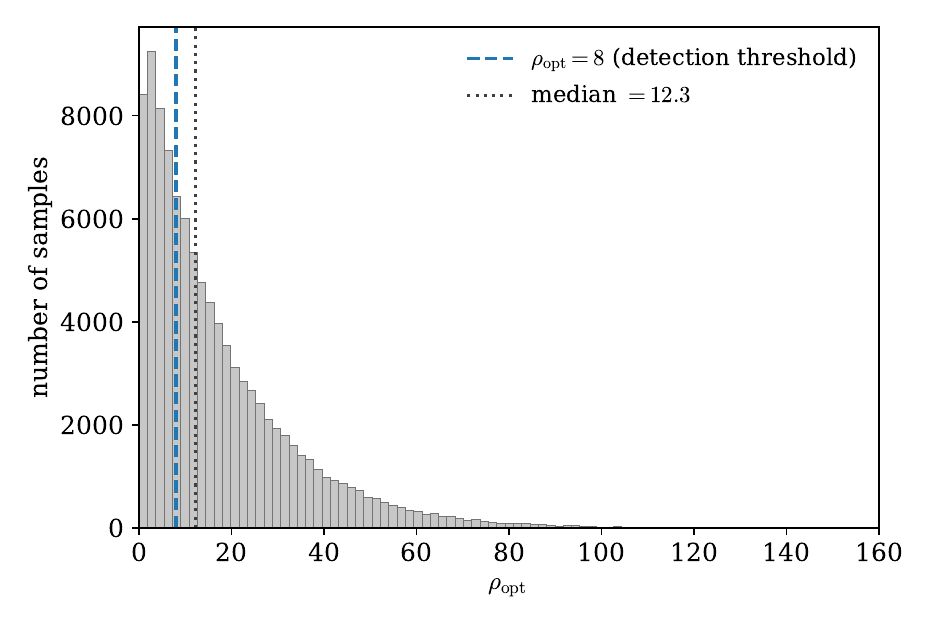}%
  }
\caption{Distribution of the training labels: the phase-averaged optimal SNR
$\rho_{\rm opt}$ at the reference distance $d_L=1000$~Mpc (shown for the
$10^5$-sample set), on a linear scale. The label is strongly right-skewed,
ranging from $\rho_{\rm opt}\lesssim0.1$ for the lightest, least favorably
oriented binaries to $\sim10^2$ for the heaviest systems, with a median of
$12.3$. This wide dynamic range motivates the $\log_{10}\rho_{\rm opt}$
regression target. The single-detector detection threshold
$\rho_{\rm opt}=8$ is shown as a dashed line. At this reference distance $\approx64\%$
of the samples exceed the detection threshold. The detectability of an astrophysical
population follows from the $\rho_{\rm opt}\propto1/d_L$ rescaling.}
\label{fig:snrdist}
\end{figure}

The distribution of the SNR labels is shown in Fig.~\ref{fig:snrdist}. At the
reference distance $d_L=1000$~Mpc the optimal SNR ranges from
$\rho_{\rm opt}\lesssim0.1$ for the lightest, least favorably oriented binaries to
$\sim10^2$ for the heaviest systems, with a median of $12.3$ and a strongly
right-skewed shape. This wide dynamic range is the primary motivation for
regressing $y=\log_{10}(\rho_{\rm opt}+10^{-12})$ rather than $\rho_{\rm opt}$
directly: it compresses the target to a comparable scale across the mass range and
prevents the loud, heavy-mass systems from dominating the loss. Because the labels
are evaluated at the fixed reference distance, the fraction exceeding the threshold
there ($\approx64\%$ at $\rho_{\rm opt}>8$) characterizes the reference
configuration rather than a population detection rate. The latter follows from the
$\rho_{\rm opt}\propto1/d_L$ rescaling.

\begin{table}[t]
\centering
\caption{Parameter sampling for the training dataset. Component masses are drawn
uniformly and reordered to $m_1\ge m_2$. The chirp mass $\mathcal{M}_c$ and mass
ratio $q$ used as features are derived from each pair. Spins and extrinsic angles
are drawn uniformly in the listed variables.}
\label{tab:sampling}
\begin{tabular}{llll}
\toprule
Parameter & Symbol & Distribution & Range \\
\midrule
Component masses        & $m_1,m_2$              & uniform ($m_1\!\ge\!m_2$) & $[3,170]\,M_\odot$ \\
\;\; (derived)          & $\mathcal{M}_c,\,q$    & ---                        & $q\in[1,170/3]$ \\
Spin magnitudes         & $|\bm{s}_{1,2}|$       & uniform                    & $[0,0.99]$ \\
Spin polar angles       & $\theta_{1,2}$         & uniform                    & $[0,\pi]$ \\
Spin azimuthal angles   & $\phi_{1,2}$           & uniform                    & $[0,2\pi]$ \\
Inclination             & $\cos\iota$            & uniform                    & $[-1,1]$ \\
Right ascension         & $\alpha$               & uniform                    & $[0,2\pi]$ \\
Declination             & $\sin\delta$           & uniform                    & $[-1,1]$ \\
Polarization            & $\psi$                 & uniform                    & $[0,\pi]$ \\
Coalescence phase       & $\phi_c$               & marginalized ($K{=}7$ avg.)& $[0,2\pi)$ \\
Luminosity distance     & $d_L$                  & fixed                      & $1000$~Mpc \\
Coalescence time        & $t_c$                  & fixed                      & $0$ \\
\bottomrule
\end{tabular}
\end{table}

\subsection{Waveform Generation and SNR Computation}
\label{sec:waveform}

For each parameter draw, we generate the plus ($h_+$) and cross ($h_\times$) polarizations using the
IMRPhenomXPHM approximant \cite{pratten2020phenomx, pratten2021phenomxhm}, which includes
inspiral-merger-ringdown dynamics, generic spin-precession effects, and higher-order modes. The training labels
therefore inherit the fidelity of this approximant. Our uniform mass sampling reaches
$q\!\approx\!57$, in the range that IMRPhenomXHM/XPHM bridges between its numerical-relativity
calibration ($q\!\le\!18$) and its Teukolsky calibration ($q\!\ge\!200$). Labels in this high-$q$
corner and at large in-plane spin, where the twisting-up precession description is least
constrained, carry the model's larger systematic uncertainty. Since the surrogate is trained to
reproduce the approximant rather than nature, this concerns the physical interpretation of the
labels, not the emulation accuracy we report. Waveforms
are computed in the frequency domain up to $f_\mathrm{Nyquist} = 1024$ Hz, padded or cropped to
match the PSD length. At this Nyquist frequency the ringdown of the lightest systems
(total mass $\sim$few~$M_\odot$, whose quasi-normal frequencies approach $\sim$kHz) is mildly
truncated. The associated SNR loss is negligible because the detector PSD is poor in that band, but it
can be removed by raising the sampling rate to 4096 Hz.

The projected detector strain is $h = F_+ h_+ + F_\times h_\times$, where antenna patterns
depend on sky location, polarization, and coalescence time through the
detector orientation. Here $F_+,F_\times$ are treated as constant over the (short, $\lesssim$ minutes)
in-band signal, the standard static-detector approximation, which is well satisfied for the
second-generation BBH regime considered here. To mitigate phase dependence, we average the SNR over
$K=7$ uniformly spaced coalescence phases $\phi_c \in [0, 2\pi)$, which equals the
zeroth Fourier coefficient ($a_0$) of the sampled sequence. This identity is exact for $\rho^2$,
a finite trigonometric polynomial in $\phi_c$. For $\rho$ itself it is an approximation, but the
$K=7$ average reproduces a finely sampled ($K=64$) phase average to better than $0.2\%$
\cite{fairhurst2010localizing}. The optimal (single-detector) SNR,
Eq.~(\ref{eq:snr}), is then computed via PyCBC's $\sigma^2$ (sigmasq) function, with fallback
to explicit summation $4\sum |\tilde h|^2/S_n\,\Delta f$ over finite, positive-PSD bins for
robustness against non-finite values.

The NN surrogate regresses not $\rho$, but $y=\log_{10}(\rho + 10^{-12})$ for numerical stability.

Computations are parallelized across $N_c=4$--$16$ cores using Joblib, with per-process Detector
instantiation to avoid pickling issues. Dataset generation time scales approximately linearly with
$N_s$ and depends on waveform settings and available parallelism. We therefore omit a single
headline timing here and focus on the modeling and evaluation protocol for dataset sizes up to 1M
samples.

\subsection{Dataset Assembly and Validation}

The resulting dataset comprises $N_s$ samples stored as a Pandas DataFrame in pickle format, with
columns $d_L$, $\log\mathcal{M}_c$, $\log q$, $\cos\iota$, the six spin components, $F_+$,
$F_\times$, $t_c$, and $\log\rho$. Before training/evaluation splits, we apply an upper-SNR
truncation and retain only samples with $\rho<200$ (i.e., $\log_{10}\rho < \log_{10}200$), as
described in Section~\ref{sec:training}. We visualize summary distributions post-generation to
confirm that the sampling covers the intended parameter ranges and does not exhibit pathological
clustering. The PyCBC-computed SNR values serve as reference labels for neural network training, and
we validate consistency by spot-checking predictions against direct PyCBC evaluations on held-out
samples.

This approach draws inspiration from prior SNR interpolation efforts
\cite{wong2020gravitational, gerosa2020gravitational} but extends to full precessing spins and
phase-averaging for increased astrophysical realism under a fixed analysis configuration. The
resulting surrogate is intended primarily for high-throughput simulation-driven workflows (e.g.,
injection campaigns for selection effects), rather than end-to-end low-latency inference from
detector strain. The single-detector restriction is not a fundamental limitation:
the network optimal SNR combines the per-detector values exactly in quadrature,
$\rho_{\rm net}=\sqrt{\sum_d \rho_d^2}$ (there are no inter-detector cross terms), so a network
selection function can be assembled by training one surrogate per detector, each with its own PSD
and antenna pattern evaluated at a common GPS time, and summing. This static-detector,
quadrature-sum construction holds for the short BBH signals studied here but breaks for hour-to-day
duration signals (low-mass binaries in ET/CE), where the antenna response varies over the signal and
must be treated as time-dependent. Future extensions could incorporate multi-detector networks and
time-varying PSDs \cite{mould2024calibrating}.

Rather than traditional $k$-fold cross-validation, which is impractical at this
training scale, we assess robustness to the dataset realization directly: all headline results use a
single generation seed (42) and a fixed training/split seed (1337) for reproducibility, and we
separately verify that the absolute accuracy and the architecture ranking are stable across two
additional, independently generated datasets (generation seeds 43 and 44,
Sec.~\ref{sec:architecture_comparison}).

\section{Classic Deep Learning Architectures}
\label{sec:classic}
This section introduces the three classic deep-learning architectures used in our SNR-regression
study (to be compared to our novel architecture we introduce in Sec.~\ref{sec:phasmanet}),
progressing from simple baselines to more structured inductive biases. Sec.~\ref{sec:mlp} describes a
multilayer perceptron that serves as a lightweight reference model for mapping the
11-dimensional parameter vector $\bm \theta$ to $\log_{10}\rho$. Sec.~\ref{sec:resnet} then considers
a 1D ResNet variant, which adds hierarchical feature extraction via convolutional residual blocks.
Finally, Sec.~\ref{sec:transformer} presents our Transformer-based regressor, leveraging
self-attention to capture global, high-order interactions between source parameters. Together, these
models provide complementary trade-offs in capacity, generalization, and computational cost that we
evaluate in the subsequent results.

\subsection{Multilayer Perceptron}
\label{sec:mlp}
Multilayer Perceptrons, foundational feed-forward networks composed of stacked linear layers
interleaved with nonlinear activations, underpin much of modern supervised learning due to their
universal approximation capabilities for continuous functions on compact domains
\cite{hornik1989multilayer}. Their simplicity, which lacks strong inductive biases such as locality or translation invariance, renders them particularly suited to {\it tabular data}, where raw features
(e.g., binary masses and spins) require no spatial preprocessing. This allows rapid prototyping and
facilitates interpretability via gradient-based attribution.

Our \texttt{MLPRegressor} uses a shallow four-hidden-layer
architecture to predict $\log_{10} \rho$ from the 11-feature parameter vector $\bm \theta$. Inputs
are mapped through four width-256 hidden layers: $11 \to 256 \to 256 \to 256 \to 256 \to 1$, with GELU activations and 0.1 dropout
per layer for regularization, mitigating overfitting. The network comprises $\sim$200k parameters, a deliberately modest capacity for
this low-dimensional regression, where larger networks tend to overfit. Each hidden layer applies a linear map followed by a GELU nonlinearity and dropout, with no
internal feature normalization. Weights are Xavier-uniform initialized and optimization is performed
with AdamW (weight decay $10^{-4}$). A summary of the main architectural choices is shown in
Table~\ref{tab:mlp_arch}.

\begin{table}[t]
\centering
\caption{\texttt{MLPRegressor} (width $w{=}256$, depth $d{=}4$) used for SNR regression. Here $B$
denotes batch size. Each hidden layer applies Linear $\rightarrow$ GELU $\rightarrow$ Dropout
($p{=}0.1$).}
\label{tab:mlp_arch}
\begin{tabular}{@{}llll@{}}
\toprule
Layer & Operation & Output shape & Notes\\
\midrule
Input  & --                   & $[B,11]$   & standardized parameter vector $\bm\theta$\\
1      & Linear $11\to256$    & $[B,256]$  & GELU $\rightarrow$ Dropout ($p{=}0.1$)\\
2      & Linear $256\to256$   & $[B,256]$  & GELU $\rightarrow$ Dropout ($p{=}0.1$)\\
3      & Linear $256\to256$   & $[B,256]$  & GELU $\rightarrow$ Dropout ($p{=}0.1$)\\
4      & Linear $256\to256$   & $[B,256]$  & GELU $\rightarrow$ Dropout ($p{=}0.1$)\\
Output & Linear $256\to1$     & $[B,1]$    & predicts $\log_{10}\hat{\rho}$\\
\bottomrule
\end{tabular}
\end{table}

Batch normalization is omitted, as is common for shallow MLPs on tabular inputs. Training
employs Huber loss ($\delta = 0.03$) with histogram-based sample weighting over 150 epochs, using
early stopping and plateau-reduced learning rates to stabilize convergence.

\subsection{ResNet}
\label{sec:resnet}
Residual Networks (ResNets), introduced in the landmark work of He et al.\ \cite{he2016deep},
transformed deep learning by addressing the degradation problem in very deep architectures through the
use of identity-based shortcut (skip) connections. In the original formulation, let $H(\boldsymbol{x})$
denote the (unknown) target mapping a stack of layers would ideally implement. ResNets instead
parameterize a residual function $F(\boldsymbol{x}) := H(\boldsymbol{x})-\boldsymbol{x}$ and output
$\boldsymbol{y}=\boldsymbol{x}+F(\boldsymbol{x})$, so each block learns a correction to the identity. This
reparameterization stabilizes gradient propagation and facilitates the training of very deep networks
with a large number of layers. Their clean identity-mapping formulation acts as an implicit regularizer
and has led to broad adoption across scientific domains, whenever hierarchical feature extraction is
beneficial \cite{he2016identity, veit2016residual}.

In gravitational-wave (GW) data analysis, ResNets are particularly effective for processing noisy
strain time series and for tasks such as event detection and denoising, where 1D convolutions
naturally extract multi-resolution waveform motifs \cite{schafer2022deep, moreno2023detecting}. For
SNR interpolation on tabular parameter inputs, ResNets offer a natural step up in structure from plain
MLPs: by treating the ordered physical parameters as a 1D sequence, convolutions can capture local
interactions (e.g., within mass or spin subsets) while depth and downsampling build progressively
more global dependencies.

Our model, \texttt{ResNetRegressor}, adapts a ResNet-18-style backbone to 1D tabular regression over
GW parameters. The 11-dimensional input vector $\bm\theta$ is treated as a univariate sequence of
shape $[B,1,11]$ (where $B$ is the batch size), in the fixed feature order of
Section~\ref{sec:methodology}, in which the mass and spin parameters each occupy contiguous positions
that short convolutional kernels can exploit. The network begins with a lightweight stem consisting of a 3$\times$1 convolution
(64 channels, stride 1, padding 1), followed by batch normalization and a ReLU nonlinearity,
producing a feature map of shape $[B,64,11]$.

The residual core comprises four stages with two residual blocks (\texttt{BasicBlock1D}) per stage
($N_B=2$). Each block contains two 3$\times$1 convolutions with batch normalization and ReLU,
combined with identity or projection shortcuts (1$\times$1 convolution + batch normalization) when
resolution or channel count changes. Downsampling is implemented via stride-2 convolutions in the
first block of stages 2--4, with channel dimensions progressing as 64 $\to$ 128 $\to$ 256 $\to$ 512.

After the final stage, adaptive average pooling aggregates over the length dimension, yielding
$[B,512]$. A small fully connected regressor then maps 512 $\to$ 256 $\to$ 1 with GELU activation
and dropout ($p=0.1$) to predict $\hat{y}=\log_{10}\hat{\rho}$. As in standard ResNets, we use
\texttt{BasicBlock1D} (no bottlenecks) to keep the model compact for short 1D inputs.
Convolutional weights use Kaiming-normal (He) initialization in fan-out mode with
a ReLU gain, batch-normalization layers are initialized to unit scale and zero shift, and the
fully-connected regression head uses PyTorch's default initialization. A summary of
the main architectural choices is shown in Table~\ref{tab:resnet_arch}.

\begin{table}[t]
\centering
\caption{\texttt{ResNet1DRegressor} architecture (1D ResNet-18-ish, 2 blocks per stage) used for SNR
regression. Shapes are shown as $[B,C,L]$ until pooling. After pooling we use $[B,D]$. The 11D
feature vector is reshaped to a length-11 1D input with one channel. Downsampling is implemented via
stride-2 convolutions in the first block of stages 2--4.}
\label{tab:resnet_arch}
\small
\setlength{\tabcolsep}{4pt}
\resizebox{\columnwidth}{!}{%
\begin{tabular}{@{}llll@{}}
\toprule
Component & Operation & Output shape & Notes\\
\midrule
Input     & reshape                                             & $[B,1,11]$   & 11D feature vector $\bm\theta$\\
Stem      & Conv $3\times1$, $C{=}64$, stride 1 $\to$ BN $\to$ ReLU & $[B,64,11]$  & padding 1\\
Stage 1   & $2\times$ \texttt{BasicBlock1D}, $C{=}64$          & $[B,64,11]$  & stride 1\\
Stage 2   & $2\times$ \texttt{BasicBlock1D}, $C{=}128$         & $[B,128,6]$  & stride-2 in first block\\
Stage 3   & $2\times$ \texttt{BasicBlock1D}, $C{=}256$         & $[B,256,3]$  & stride-2 in first block\\
Stage 4   & $2\times$ \texttt{BasicBlock1D}, $C{=}512$         & $[B,512,2]$  & stride-2 in first block\\
Pooling   & AdaptiveAvgPool1d(1)                               & $[B,512,1]$  & global average over $L$\\
Flatten   & squeeze                                            & $[B,512]$    & remove length dim\\
Regressor & Linear $512\to256\to1$                             & $[B,1]$      & GELU + Dropout ($p{=}0.1$)\\
\bottomrule
\end{tabular}%
}
\end{table}

Relative to conventional 2D ResNets, our implementation makes four changes.
First, it uses 1D convolutions along the parameter axis (e.g., kernels sliding over spin triplets to
exploit local vectorial structure).
Second, it down-samples primarily via strides, avoiding aggressive pooling that would collapse the already
short sequence.
Third, it replaces the final softmax classifier with a regression output.
Fourth, it operates on a fixed-length 11-feature input sequence (no variable-length masking).

With roughly 3.98M parameters and $\sim$2 ms inference time per batch on GPU, the model is suitable
for high-throughput inference in simulation-driven workflows \cite{wong2020gravitational,
magee2024neural}. In our experiments, residual (skip) connections improve optimization stability and
validation convergence relative to a plain 1D CNN variant under the same training protocol,
consistent with prior findings on residual learning in deep networks.

The main architectural hyperparameters follow a standard ResNet-18
design and are kept fixed rather than tuned for this task: block counts (two per stage), kernel sizes
(3$\times$1), dropout (0.1, in the regression head), and channel widths. The shared training hyperparameters are
inherited from the MLP baseline (Section~\ref{sec:training}). Under the shared training configuration (Huber loss with
$\delta = 0.03$ and histogram-based sample weighting, as described in Section~\ref{sec:training}),
the model serves as a competitive convolutional baseline for this structured regression task,
benefiting from residual connections and a local inductive bias along the feature axis.

\subsection{Transformer}
\label{sec:transformer}

Transformers constitute a major shift in deep learning, originally introduced to replace recurrent
neural networks (RNNs) in sequence modeling through the mechanism of scaled dot-product self-attention
\cite{vaswani2017attention}. By computing pairwise interactions between all tokens in parallel,
Transformers capture long-range dependencies without the sequential bottlenecks inherent to RNNs,
enabling efficient training on large datasets and long contexts. Following their widespread success in
natural language processing (powering models such as BERT and GPT) Transformer architectures have
been adapted to computer vision (e.g., Vision Transformers) and even to tabular data regression
\cite{dosovitskiy2020image, gorishniy2021revisiting}.

In our setting, a Transformer encoder provides a convenient attention-based baseline for modeling
non-linear interactions among the 11 input features. Self-attention can represent feature--feature
couplings without imposing locality assumptions, making it a reasonable comparator for structured
regression in this fixed-size tabular surrogate task. We therefore include a tailored
\texttt{TransformerRegressor} as a strong attention baseline under the same training protocol.

Our architecture, \texttt{TransformerRegressor}, is a tailored Transformer encoder optimized for SNR
regression on fixed-size tabular GW data. Unlike Transformers designed for variable-length sequences,
our model fixes the sequence length to $S = 11$ to match the dimensionality of the physical parameter
vector $\bm \theta$, avoiding padding and improving computational efficiency. Each scalar
feature, including intrinsic parameters ($\log_{10}\mathcal{M}_c$, $\log_{10} q$), spin components
($\boldsymbol{s}_1, \boldsymbol{s}_2$), and extrinsic parameters ($\cos\iota$, $F_+$, $F_\times$,
$d_L$), is embedded into a $d_\mathrm{model}=128$-dimensional token via a linear projection (input
dim: 1 $\to$ 128).

To encode feature identity/ordering information and break permutation symmetry (e.g., distinguishing
primary vs.\ secondary spin vectors), we employ learnable positional embeddings of shape
$[1, S, d_\mathrm{model}]$, initialized with small Gaussian noise ($\mu=0$, $\sigma=0.02$). The
encoder comprises $N_L = 4$ layers, each containing multi-head self-attention with $N_h = 8$ heads
and a position-wise feed-forward sublayer with $d_\mathrm{ff} = 256$ and a GELU activation. We use
pre-layer normalization to enhance training stability, especially important in deeper stacks
\cite{ba2016layer}. Dropout ($p = 0.1$) is applied within both the attention and MLP blocks to
reduce overfitting. The encoder output of shape $[B, S, d_\mathrm{model}]$ is aggregated using
global mean pooling over the sequence dimension, followed by a LayerNorm, yielding a single
representation $[B, d_\mathrm{model}]$ more suitable for tabular inputs than CLS-token approaches.
The pooled representation feeds a lightweight regression head consisting of two linear layers
(128 $\to$ 64 $\to$ 1) with GELU and dropout. Token projection layers use Xavier-uniform
initialization to maintain stable variance propagation. A summary of the main architectural choices
is shown in Table~\ref{tab:transformer_arch}.

\begin{table}[t]
\centering
\caption{\texttt{TransformerRegressor} architecture used for SNR regression. Shapes are shown as
$[B,S,d]$ for batch size $B$, sequence length $S{=}11$, and embedding dimension
$d=d_\mathrm{model}=128$.}
\label{tab:transformer_arch}
\small
\setlength{\tabcolsep}{4pt}
\begin{tabular}{@{}lll p{0.46\columnwidth}@{}}
\toprule
Component      & Operation                            & Output shape & Notes\\
\midrule
Input          & --                                   & $[B,11]$     & parameter vector $\bm\theta$\\
Reshape        & $\bm\theta \mapsto \bm\theta[:,:,None]$ & $[B,11,1]$ & treat each feature as a token\\
Tokenization   & Linear $1\to128$ (per token)         & $[B,11,128]$ & scalar $\to$ token embedding\\
Position enc.  & Learnable embeddings                 & $[B,11,128]$ & added to tokens\\
Encoder        & $N_L{=}4$ Transformer layers         & $[B,11,128]$ & pre-LN; MHA: $N_h{=}8$;
                                                                         FFN: $d_\mathrm{ff}{=}256$;
                                                                         GELU; dropout $p{=}0.1$\\
Pooling        & Mean over $S$ $\to$ LayerNorm        & $[B,128]$    & global mean pooling + LN\\
Regressor      & Linear $128\to64\to1$                & $[B,1]$      & GELU + dropout ($p{=}0.1$)\\
\bottomrule
\end{tabular}
\end{table}

Compared to the standard Transformer encoder, our model differs in three key respects.
First, it uses no decoder, as the task is direct regression.
Second, it applies mean pooling instead of CLS-token aggregation, promoting more holistic feature fusion.
Third, it uses a batch-first layout for seamless PyTorch integration.
The model is lightweight and computationally efficient for fixed-length tabular regression. We use it
as a strong attention-based baseline under the same data splits and training protocol used for the
other architectures employed in this study.

The training hyperparameters (learning rate $\eta = 10^{-3}$, batch size $B = 128$, maximum epochs
150, early-stopping patience 20) are the shared settings tuned on the MLP baseline (Section~\ref{sec:training})
and applied here unchanged rather than re-tuned for this model. We train
using Huber loss with $\delta = 0.03$ in $\log_{10}\rho$ space, together with the same
histogram-based sample weighting scheme described in Section~\ref{sec:training}. This weighting is
computed on the training split only, clipped to $[0.2,5.0]$, and renormalized to mean 1.0 for
numerical stability, and is applied uniformly across all architectures.

\subsection{Post-processing}
As a post-processing step, common to all architectures we implement, we apply a monotonic, shape-preserving {\it calibrator} based on a
monotonic Piecewise Cubic Hermite Interpolation (PCHIP) fitted on the validation set and applied to
test predictions. This mapping can reduce global systematic bias without changing the rank order of
predictions, providing a simple correction that is computationally negligible at inference time. We
report results both before and after this calibration procedure.

\section{\textsc{$\Phi$asmaNet}: A Staged Spectral Token-Refinement Network with Recursive Feedback}
\label{sec:phasmanet}

\noindent This section introduces \textsc{$\Phi$asmaNet}, our novel architecture for fast, accurate
SNR regression from low-dimensional GW source parameters. The key idea is to represent the 11 input features as a small set of learned tokens
and refine them progressively through a short, ordered sequence of $\Phi$-Stages, each combining a standard Transformer encoder layer with the novel $\Phi$-Refractor
(spectral subspace mixing module), augmented with a small number of $\Phi$-Feedback (gated top-down
feedback) iterations. This design aims to combine the global interaction modeling of Transformers
with the hierarchical inductive bias of staged computation, while remaining efficient on fixed-length
tabular inputs.

\begin{figure}[t]
  \centering
  \makebox[\textwidth][c]{%
    \includegraphics[
      width=1\textwidth,
      height=0.5\textheight,
      keepaspectratio
    ]{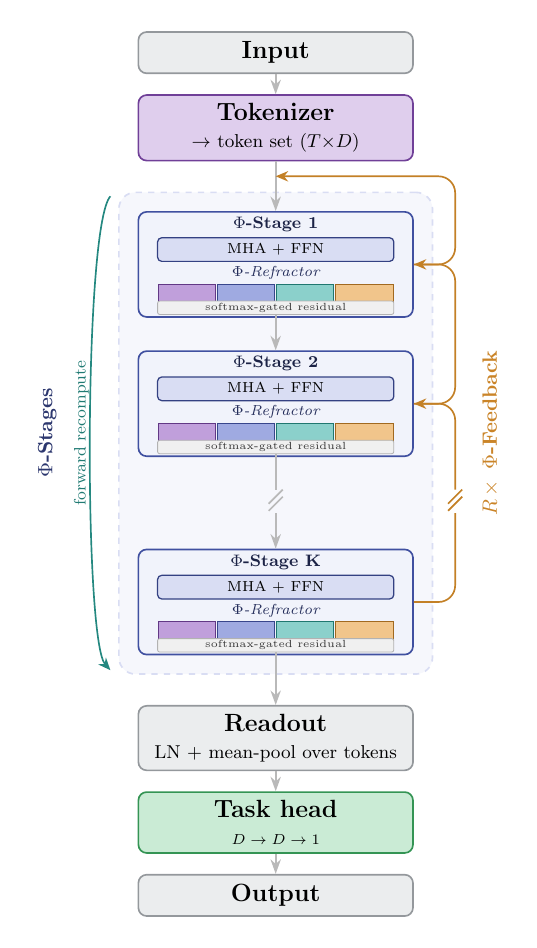}%
  }
  \caption{\textbf{\textsc{$\Phi$asmaNet} overview.} A fixed-dimensional structured input is tokenized into a learned token
  set of shape $(T\times D)$ and processed by an ordered sequence of $K$ $\Phi$-Stages (each
  $\Phi$-Stage: standard Transformer encoder layer + $\Phi$-Refractor). The model then performs $R$
  $\Phi$-Feedback passes (top-down gated feedback with forward recomputation), before token pooling
  (readout) and a task-specific head produce the final output.}
  \label{fig:cascade_forge_arch}
\end{figure}

As shown in Figure~\ref{fig:cascade_forge_arch}, \textsc{$\Phi$asmaNet} implements a
staged-refinement computation pattern tailored to short, fixed-length structured inputs. The network
begins with a linear \emph{tokenizer} that maps the input feature vector to a compact token set of
size $T$ with embedding dimension $D$. These tokens are
then processed by an \emph{ordered} sequence of $K$ $\Phi$-Stages, where each $\Phi$-Stage combines
a standard Transformer encoder layer (multi-head self-attention for token--token interaction followed
by an FFN, with residual connections and normalization) and the $\Phi$-Refractor that re-mixes
embedding subspaces via parallel subspace transforms and learned gating. After the initial forward pass through the ordered $\Phi$-Stages reaches the top level, 
the model unrolls $R$ $\Phi$-Feedback iterations: a top-down gated feedback
pass injects higher-level context into lower levels, followed by a forward recomputation through all
$\Phi$-Stages to propagate the updates consistently. Finally, a \emph{readout} module aggregates the
refined tokens (LayerNorm + mean pooling in our default configuration) to produce a single
representation, which is passed to a lightweight task head (e.g., a small MLP regressor) to output
the prediction (here, $\log_{10}(\rho)$ for SNR regression).

\subsection{Tokenization: From Tabular Parameters to Pseudo-Sequence}
The tokenization module forms \textsc{$\Phi$asmaNet}'s entry point, translating 
11-dimensional BBH parameters $\bm\theta$ into a compact pseudo-sequence of $T=8$
learnable tokens embedded in $D=128$ dimensions:
\begin{equation}
\boldsymbol{E}_0=\boldsymbol{W}_T \boldsymbol{\theta}+\boldsymbol{b}_T \in \mathbb{R}^{B \times T \times D}
\end{equation}
with $\boldsymbol{W}_T  \in \mathbb{R}^{TD \times 11}$ initialized via 
PyTorch's default Kaiming-uniform scheme. This projection distributes features across tokens and can encourage
specialization across the token set. We do not explicitly interpret individual tokens.

Converting $\bm\theta$ into a short sequence enables self-attention to capture 
non-local dependencies among physical parameters without requiring hand-crafted 
feature interactions. No explicit positional encodings are used. The network relies
on the tokenizer projection and the learned attention weights to distinguish feature 
roles. We treat this tokenization as a design choice for compactness. Isolating its
marginal impact would require additional ablations beyond the scope of this manuscript.

\subsection{$\Phi$-Stages: Progressive Abstraction via Attention, FFN, and Spectral Mixing}

\textsc{$\Phi$asmaNet} stacks $K=3$ $\Phi$-Stages to transform raw tokens into hierarchical
embeddings. Let $\operatorname{LN}$ denote LayerNorm, $\operatorname{MHA}$ multi-head
self-attention with $N_h=4$ heads, $\operatorname{GELU}$ the GELU activation applied
elementwise, and $\mathcal{D}_p$ dropout with probability $p=0.1$. Each $\Phi$-Stage
consists of two sublayers. First, a standard Transformer encoder layer processes
$\boldsymbol{x} \in \mathbb{R}^{B \times T \times D}$ via self-attention followed by a
position-wise feed-forward network (FFN), with residual connections and post-layer
normalization:
\begin{equation}
\boldsymbol{x}^{\prime}
= \operatorname{LN}\!\left(\boldsymbol{x}
   + \mathcal{D}_p\!\bigl(\operatorname{MHA}(\boldsymbol{x},\boldsymbol{x},\boldsymbol{x})\bigr)\right),
\end{equation}
\begin{equation}
\boldsymbol{x}^{\prime\prime}
= \operatorname{LN}\!\left(\boldsymbol{x}^{\prime}
   + \mathcal{D}_p\!\left(
      \boldsymbol{W}_2\,\operatorname{GELU}\!\left(\boldsymbol{W}_1 \boldsymbol{x}^{\prime}
      + \boldsymbol{b}_1\right) + \boldsymbol{b}_2
   \right)\right),
\end{equation}
where
$\boldsymbol{W}_1 \in \mathbb{R}^{4D \times D}$ and
$\boldsymbol{W}_2 \in \mathbb{R}^{D \times 4D}$
are the FFN expansion and projection weights, with biases
$\boldsymbol{b}_1 \in \mathbb{R}^{4D}$ and
$\boldsymbol{b}_2 \in \mathbb{R}^{D}$. Second, the $\Phi$-Refractor (described below)
performs multi-perspective subspace mixing on $\boldsymbol{x}^{\prime\prime}$.

Stacking $\Phi$-Stages is intended to refine representations progressively, with
earlier stages encouraged to capture more local parameter interactions (for instance among spin
components, or between mass and spin) and later stages to assemble these into the more global
structure that sets the SNR. We stress that this local-to-global ordering is the design motivation
rather than a verified property: as with the individual tokens, we do not probe what each stage
encodes, and any such specialization would emerge from training rather than being imposed. Within
each stage, residual connections and normalization support stable optimization across the stack.

\subsection{$\Phi$-Refractor: Multi-Perspective Subspace Mixing}
\label{subsec:phi-refractor}

The $\Phi$-Refractor is a switchable component within each $\Phi$-Stage that introduces
multi-perspective subspace processing. It partitions the token embeddings into $P$ parallel
subspaces, applies independent subspace-specific transformations to promote diverse feature
specializations, and fuses the resulting representations via learned gating. This design encourages
the network to explore multiple complementary ``views'' of the input, which we find beneficial for
structured low-dimensional regression tasks such as BBH SNR estimation. In this work we use
$\Phi$-Refractor \textbf{ON} as the default configuration for our primary SNR regression
experiments, and treat $\Phi$-Refractor \textbf{OFF} as a controlled architectural ablation under
an otherwise matched protocol (Section~\ref{sec:training}).

Formally, let $\boldsymbol{h} \in \mathbb{R}^{B \times T \times D}$ denote the token embedding at a
given $\Phi$-Stage, where $B$ is the batch size, $T$ the number of tokens, and $D$ the embedding
dimension. The $\Phi$-Refractor operates by partitioning the embedding dimension into $P$ disjoint
subspaces with sizes $\{D_p\}_{p=1}^{P}$ such that $\sum_{p=1}^{P} D_p = D$. In our experiments
we use equal-sized subspaces $D_p = D/P$ for all $p$. Writing the partition as a channel-wise
concatenation,
\begin{equation}
\boldsymbol{h} = \big[\,\boldsymbol{h}^{(1)} \,\Vert\, \boldsymbol{h}^{(2)} \,\Vert\, \cdots
\,\Vert\, \boldsymbol{h}^{(P)}\,\big], \qquad \boldsymbol{h}^{(p)} \in \mathbb{R}^{B \times T \times D_p},
\end{equation}
each subspace is transformed independently by a lightweight two-layer MLP $\varphi_p$:
\begin{equation}
\boldsymbol{t}^{(p)} = \varphi_p\!\left(\boldsymbol{h}^{(p)}\right), \qquad p = 1,\ldots,P,
\end{equation}
and the transformed subspaces are concatenated back into a unified representation:
\begin{equation}
\boldsymbol{t} = \big[\,\boldsymbol{t}^{(1)} \,\Vert\, \boldsymbol{t}^{(2)} \,\Vert\, \cdots
\,\Vert\, \boldsymbol{t}^{(P)}\,\big] \in \mathbb{R}^{B \times T \times D}.
\end{equation}
Feature-wise gating weights are computed via a softmax-normalized linear map applied along the
embedding dimension $D$:
\begin{equation}
\boldsymbol{w} = \mathrm{softmax}\!\left(\boldsymbol{W}_r \boldsymbol{t} + \boldsymbol{b}_r\right)
\in \mathbb{R}^{B \times T \times D},
\end{equation}
and the final output is produced by a gated residual recombination:
\begin{equation}
\operatorname{\Phi\text{-}Refractor}(\boldsymbol{h})
\;=\; \boldsymbol{t} \;+\; \mathcal{D}_p\!\bigl(\boldsymbol{w} \odot \boldsymbol{t}\bigr).
\end{equation}
where $\odot$ denotes element-wise multiplication. This design encourages specialization across
subspaces (via independent $\varphi_p$) while retaining stable optimization through residual
routing. In practice, it acts as a structured subspace mixing and gating module that improves
representational flexibility without changing the token length $T$ or the embedding dimension $D$.

On the Covertype benchmark (see Appendix~\ref{sec:appendix}), enabling the
$\Phi$-Refractor also improves accuracy. A broader ablation study isolating the marginal
contribution of each component across regimes is left for future work.

\subsection{$\Phi$-Feedback: Recursive Top-Down Gated Refinement}

\textsc{$\Phi$asmaNet} employs $R=2$ $\Phi$-Feedback passes, propagating top-down feedback via learnable
gates:
\begin{equation}
\begin{aligned}
\boldsymbol{g}_l &= \begin{cases}
\sigma(\boldsymbol{W}_g^l\,\boldsymbol{E}_l), & l = 1,\ldots,K-1,\\
\sigma(\boldsymbol{E}_l), & l = K,
\end{cases}\\
\boldsymbol{E}_{l-1} &\leftarrow \boldsymbol{E}_{l-1} + \boldsymbol{g}_l \odot \boldsymbol{E}_l, \qquad l = K,\ldots,1.
\end{aligned}
\end{equation}
where $\boldsymbol{W}_g^l$ denotes a learned linear projection mapping the level-$l$ token embeddings
$\boldsymbol{E}_l \in \mathbb{R}^{B \times T \times D}$ to gate logits for the lower
$K-1$ links, while the topmost link ($l=K$) applies a parameterless sigmoid gate
$\sigma(\boldsymbol{E}_K)$. Here $\sigma(\cdot)$ is the sigmoid
applied elementwise, so $\boldsymbol{g}_l \in [0,1]^{B \times T \times D}$. \textsc{$\Phi$asmaNet}
therefore carries $K-1$ learned gate projections, shared across the $R$ feedback passes. This top-down flow allows
global embeddings to modulate local motifs selectively. In this paper we keep $\Phi$-Feedback
enabled in \textsc{$\Phi$asmaNet}, and treat the $\Phi$-Refractor as a switchable component (default
$\Phi$-Refractor ON for SNR regression, with OFF reported as an ablation, and enabled where beneficial
for tabular classification). Isolating the marginal contribution of each component and analyzing
regime-specific effects (e.g., low-SNR subsets) are left for future work.

\subsection{Regression Head: Holistic Pooling and Output Projection}

The regression head aggregates the top-level token embeddings
$\boldsymbol{E}_K \in \mathbb{R}^{B \times T \times D}$ via per-token layer
normalization followed by mean pooling over the token axis:
\begin{equation}
\boldsymbol{z} \;=\; \frac{1}{T}\sum_{t=1}^{T}
\bigl[\operatorname{LN}(\boldsymbol{E}_K)\bigr]_{t},
\end{equation}
where $\boldsymbol{z} \in \mathbb{R}^{B \times D}$ and
$[\,\cdot\,]_{t}$ denotes the $t$-th slice along the token axis. The pooled
representation is mapped to the scalar prediction by a compact two-layer MLP:
\begin{equation}
\hat{y} \;=\; \log_{10}\hat{\rho} \;=\;
\boldsymbol{w}_o^{\!\top}\,\operatorname{GELU}\!\bigl(\boldsymbol{W}_h\,\boldsymbol{z}
+ \boldsymbol{b}_h\bigr) + b_o ,
\end{equation}
with hidden weights $\boldsymbol{W}_h \in \mathbb{R}^{D \times D}$ and
$\boldsymbol{b}_h \in \mathbb{R}^{D}$, and output weights
$\boldsymbol{w}_o \in \mathbb{R}^{D}$, $b_o \in \mathbb{R}$. Mean pooling
provides a simple and robust aggregation of token information for fixed-length
structured inputs, yielding stable performance in our setting.

\begin{algorithm}[H]
\caption{\textsc{$\Phi$asmaNet} forward pass with ordered $\Phi$-Stages and $\Phi$-Feedback.}
\label{alg:phasmanet-forward}
\begin{algorithmic}[1]
\Require Input features $x \in \mathbb{R}^{B \times F}$; tokenizer $\mathrm{Tok}$; Transformer
encoder layers $\{\mathrm{C}_i\}_{i=0}^{K-1}$; $\Phi$-Refractors $\{\mathrm{P}_i\}_{i=0}^{K-1}$;
$\Phi$-Feedback gates $\{\mathrm{G}_\ell\}_{\ell=0}^{K-2}$; recursion steps $R$
\Ensure Prediction $\hat{y} \in \mathbb{R}^{B \times 1}$ (e.g., $\log_{10}(\widehat{\rho})$)

\State $E_0 \gets \mathrm{Tok}(x)$ \Comment{$E_0 \in \mathbb{R}^{B \times T \times D}$}
\State $\mathcal{E} \gets [E_0]$ \Comment{Store all levels}

\Statex
\State \textbf{Initial forward ordered staged refinement ($K$ ordered $\Phi$-Stages)}
\For{$i \gets 0$ \textbf{to} $K-1$}
    \State $H \gets \mathrm{C}_i(\mathcal{E}[i])$
    \Comment{Standard Transformer encoder layer}
    \If{$\Phi\text{-Refractor ON}$}
        \State $H \gets \mathrm{P}_i(H)$ \Comment{Spectral subspace mixing}
    \EndIf
    \State $\mathcal{E}.\mathrm{append}(H)$
\EndFor

\Statex
\State \textbf{$\Phi$-Feedback (unrolled $R$ times)}
\For{$r \gets 1$ \textbf{to} $R$}
    \Statex \Comment{Backward (top-down) gated feedback}
    \For{$\ell \gets K$ \textbf{down to} $1$}
        \State $E_{\text{high}} \gets \mathcal{E}[\ell]$
        \State $E_{\text{low}}  \gets \mathcal{E}[\ell-1]$
        \If{$\ell-1 \le K-2$}
            \State $g \gets \mathrm{G}_{\ell-1}(E_{\text{high}})$
            \Comment{Learned gate, $g \in \mathbb{R}^{B \times T \times D}$}
        \Else
            \State $g \gets \sigma(E_{\text{high}})$ \Comment{Fallback gate on topmost link}
        \EndIf
        \State $\mathcal{E}[\ell-1] \gets E_{\text{low}} + g \odot E_{\text{high}}$
    \EndFor

    \Statex \Comment{Forward recompute through all $\Phi$-Stages}
    \For{$i \gets 0$ \textbf{to} $K-1$}
        \State $H \gets \mathrm{C}_i(\mathcal{E}[i])$
        \If{$\Phi\text{-Refractor ON}$}
            \State $H \gets \mathrm{P}_i(H)$
        \EndIf
        \State $\mathcal{E}[i+1] \gets H$
    \EndFor
\EndFor

\Statex
\State \textbf{Readout and regression head}
\State $Z \gets \mathrm{LN}(\mathcal{E}[K])$
\State $z \gets \mathrm{MeanPool}(Z, \text{tokens})$
\State $\hat{y} \gets \mathrm{Regressor}(z)$
\State \Return $\hat{y}$
\end{algorithmic}
\end{algorithm}

Algorithm~\ref{alg:phasmanet-forward} summarizes the complete \textsc{$\Phi$asmaNet} forward pass in a
single pseudocode listing. In particular, it makes explicit the bookkeeping over the hierarchy
(via the level list $\mathcal{E}$) and the unrolled schedule that alternates top-down $\Phi$-Feedback
updates with a full forward recomputation, as well as the conditional execution of the
$\Phi$-Refractor used in our ablation settings.

\subsection{Model Summary}
\label{subsec:model-summary}

Having described each component in turn, we summarize the concrete
\textsc{$\Phi$asmaNet} instantiation used throughout our SNR-regression experiments
(Table~\ref{tab:phasmanet_arch}). The $F=11$ standardized input features are mapped by the tokenizer
to a set of $T=8$ tokens of embedding dimension $D=128$. We then apply $K=3$ ordered $\Phi$-Stages,
each pairing a standard Transformer encoder layer ($N_h=4$ attention heads, FFN expansion
$d_\mathrm{ff}=4D=512$, GELU, dropout $p=0.1$) with a $\Phi$-Refractor that splits the embedding into
$P=4$ equal subspaces ($D_p=32$), transforms each by an independent two-layer MLP, and recombines
them through a softmax-gated residual. The model unrolls $R=2$ $\Phi$-Feedback passes, each a
top-down gated update ($K-1=2$ learned gate projections and a parameterless gate on the topmost link)
followed by a forward recomputation through all $\Phi$-Stages. A per-token LayerNorm and mean-pooling
readout feeds a two-layer regression head ($128\to128\to1$, GELU) that outputs $\log_{10}\hat{\rho}$.
The $\Phi$-Refractor is enabled by default and treated as a switchable ablation. The full model
totals ${\sim}732$k trainable parameters (Table~\ref{tab:param-counts}).

The $\Phi$-Refractor-OFF ablation is implemented by removing the $\Phi$-Refractor
(i) from the initial forward pass through the ordered $\Phi$-Stages and (ii) from the recomputation
step inside each $\Phi$-Feedback pass, yielding a fully matched comparison.

\begin{table}[t]
\centering
\caption{\textsc{$\Phi$asmaNet} architecture used for SNR regression. Shapes are shown as $[B,T,D]$
for batch size $B$, token count $T{=}8$, and embedding dimension $D{=}128$.}
\label{tab:phasmanet_arch}
\small
\setlength{\tabcolsep}{4pt}
\begin{tabular}{@{}lll p{0.34\columnwidth}@{}}
\toprule
Component       & Operation                          & Output shape  & Notes\\
\midrule
Input           & --                                 & $[B,11]$      & standardized parameter vector $\bm\theta$\\
Tokenizer       & Linear $11\to TD$, reshape         & $[B,8,128]$   & learned token set $\boldsymbol{E}_0$\\
$\Phi$-Stage ($\times3$)    & Enc.\ layer $+$ $\Phi$-Refractor   & $[B,8,128]$   & MHA $N_h{=}4$, FFN $d_\mathrm{ff}{=}512$, GELU; $\Phi$-Refractor $P{=}4$ ($D_p{=}32$), softmax-gated residual; dropout $0.1$\\
$\Phi$-Feedback ($\times2$) & Gated update $+$ recompute         & $[B,8,128]$   & $K{-}1{=}2$ learned gates $+$ parameterless top gate\\
Readout         & Per-token LN $\to$ mean over $T$   & $[B,128]$     & global mean pooling\\
Reg.\ head      & Linear $128\to128\to1$             & $[B,1]$       & GELU; outputs $\log_{10}\hat{\rho}$\\
\bottomrule
\end{tabular}
\end{table}

\section{Architectural Novelty and Design Rationale}
\label{sec:novelty}

Most contemporary deep learning architectures are designed around a single dominant paradigm:
Transformers apply self-attention uniformly across depth, residual networks rely on identity-based
skip connections without enforcing semantic differentiation across layers, and MLPs apply
homogeneous transformations differing only in width or depth. For structured regression tasks such
as SNR estimation, where representations may benefit from being refined incrementally from local signal
characteristics to stable global descriptors, {\it these uniform-depth assumptions may not provide the
most effective inductive bias}.

\textsc{$\Phi$asmaNet} addresses this by organizing computation into an {\it ordered
sequence of $\Phi$-Stages} coupled by an explicit top-down pathway. Unlike a standard Transformer
encoder, whose layers are architecturally symmetric by design and provide hierarchy only through depth, \textsc{$\Phi$asmaNet}'s stages are non-exchangeable: the $\Phi$-Feedback mechanism propagates
context from higher to lower stages, so that the stage \emph{order} is architecturally meaningful,
while the $\Phi$-Refractor adds multi-perspective subspace diversification after each attention layer.
This explicit top-down refinement pathway is {\it absent from conventional Transformer, ResNet, and
MLP designs}. The staged organization is \emph{intended} to encourage a directional, local-to-global
refinement of the representation. As discussed in Section~\ref{sec:phasmanet}, we treat this division
of labor as the design motivation rather than a demonstrated property, and defer its empirical
verification (e.g., reorder/collapse ablations) to future work.

For our current application, under identical training and evaluation conditions, \textsc{$\Phi$asmaNet} outperforms the evaluated
Transformer-, ResNet-, and MLP-based neural baselines across the dataset scales considered,
consistent with the staged organization acting as a useful inductive bias for this surrogate task.
While \textsc{$\Phi$asmaNet} is evaluated here in the context of SNR estimation, its architectural
principles are not inherently domain-specific and may be applicable to other structured
tabular regression and classification tasks that benefit from staged, hierarchical
refinement, as the out-of-the-box benchmarks of Appendix~\ref{sec:appendix} preliminarily indicate.

We emphasize that the contribution of \textsc{$\Phi$asmaNet} lies in architectural
organization and inductive-bias enforcement, that is, in how standard components are composed and trained
together to impose a non-homogeneous, staged bias, rather than in proposing new neural primitives
or optimization techniques.

We note that the $\Phi$-Refractor is mechanistically related to several established design patterns: its partition of the embedding dimension into disjoint subgroups with independent per-group transformations follows the split-transform-merge paradigm introduced by ResNeXt \cite{xie2017aggregated} and, more broadly, grouped processing designs. Its learned softmax-weighted recombination shares the spirit of channel gating mechanisms such as Squeeze-and-Excitation \cite{hu2018squeeze}. The $\Phi$-Refractor also differs from multi-head self-attention, which splits the embedding into subspaces but applies dot-product attention across tokens within each head and recombines by concatenation followed by a linear projection \cite{vaswani2017attention}. By contrast, the $\Phi$-Refractor applies independent position-wise MLPs within each subspace and recombines via element-wise softmax-weighted gating with a residual connection. To our knowledge, the specific combination of these elements, namely disjoint embedding-dimension partitioning, independent subspace-specific MLP transforms, and softmax-gated residual recombination, applied as a post-attention module within each $\Phi$-Stage, has not appeared in prior work in this form.

Note also that the `spectral' and `refractor' terminology is used as an architectural metaphor for subspace diversification and does not imply spectral decomposition in the signal-processing or linear-algebraic sense. The design is motivated by the physical heterogeneity of the input parameter space: the 11-dimensional feature vector comprises quantities from distinct physical domains (masses governing waveform morphology, spin vectors encoding precession dynamics, and antenna-pattern projections capturing geometric orientation effects), which enter the SNR through qualitatively different coupling mechanisms. Although the $\Phi$-Refractor partitions the learned embedding dimension rather than the raw input features, and therefore does not enforce an explicit alignment between subspaces and physical parameter groups, the intent is to provide the network with a structural incentive for developing multiple complementary processing pathways within each $\Phi$-Stage, that may reflect this underlying physical heterogeneity. Whether the learned subspaces develop interpretable physical specialization is an empirical question that we leave for future investigation.

\section{Training Protocol, Hyperparameters, and Evaluation Metrics}
\label{sec:training}

All models are trained under a matched protocol: identical data split, preprocessing, loss
definition, sample weighting, optimizer, learning-rate schedule, and early stopping.
Architecture-specific internal regularization (e.g., dropout) follows each model's fixed
configuration. The key shared settings are summarized below, while full details appear in the
remainder of this section. 

Our primary goal is a controlled comparison among neural surrogates that share a matched optimization
protocol and are differentiable end-to-end, allowing them to be embedded directly into larger differentiable analysis pipelines. We therefore compare
against three representative neural regressors (MLP, 1D-ResNet, Transformer encoder) trained under
identical splits, preprocessing, loss, scheduling, and early stopping. We note that strong non-neural
tabular baselines such as gradient-boosted decision trees (e.g., LightGBM/CatBoost) can perform
competitively on low-dimensional structured data. An explicit comparison to such methods is an
important extension and is left for future work.

\paragraph{Per-architecture tuning.} The shared training
hyperparameters (optimizer, learning rate, weight decay, schedule, batch size, loss, weighting, and
early stopping) were tuned on the MLP baseline and then held fixed and unchanged for the
1D-ResNet, the Transformer encoder, and \textsc{$\Phi$asmaNet}, each of which is used at a fixed
architectural configuration (for \textsc{$\Phi$asmaNet}, $T{=}8$, $D{=}128$, $K{=}3$, $R{=}2$,
$P{=}4$, inherited unchanged for the cross-domain benchmarks of Appendix~\ref{sec:appendix}) with no
per-model hyperparameter optimization. \textsc{$\Phi$asmaNet}'s advantage is therefore obtained under
optimization settings selected for a different, simpler model, with no tuning carried out in its
favor.


Before splitting the data, we retain samples with $\rho < 200$ and discard events with $\rho \ge
200$. We treat this as a \emph{definition of the target operating distribution} for the surrogate
model studied here: it removes extremely loud outliers that can dominate weighted regression
objectives and allows us to focus evaluation on the bulk of the simulated SNR range under the
fixed-PSD, single-detector setup. This filter is applied uniformly to all architectures and to all
splits. All reported dataset sizes refer to the filtered distribution unless explicitly stated
otherwise.

The dataset was partitioned into training, validation, and held-out test subsets in proportions of
70\%, 15\%, and 15\%, respectively. All splits (70/15/15) are applied after preprocessing (including
the upper-SNR cut). Therefore the absolute number of samples in each split reflects the effective
dataset size after filtering.

To facilitate effective learning, all input features are standardised to zero mean and unit variance,
using mean and variance computed exclusively from the training split and then applied to the
validation and test splits to prevent data leakage. The target variable, $\log_{10}\rho_{\rm opt}$,
is left in its natural logarithmic scale to preserve interpretability and numerical stability.
Training was conducted in mini-batches of 128 samples, over a maximum of 150 epochs. Early stopping
was employed based on the validation Huber loss, with a patience of 20 epochs, to prevent overfitting
while allowing sufficient opportunity for convergence. Dropout of 0.1 was applied wherever applicable
to further regularise the models.

Optimisation was performed using {\tt AdamW} with a learning rate of $10^{-3}$ and a weight
decay coefficient of $10^{-4}$ for {all} architectures (MLP, ResNet, Transformer, and
\textsc{$\Phi$asmaNet}). In all cases, a {\tt ReduceLROnPlateau} scheduler reduced the learning rate by a
factor of $0.5$ if the validation loss did not improve for five consecutive epochs. The {\it Huber}
loss function with $\delta = 0.03$ in logarithmic $\rho$ space was used to balance robustness to
outliers with sensitivity to small deviations.

To mitigate imbalance across the target distribution, we use a histogram-based weighting scheme
computed on the training split only (\texttt{by\_hist}). We bin the training targets $\log_{10}\rho$
into fixed bins and assign each sample a weight inversely proportional to the empirical bin
frequency. We clip weights to $[0.2,5.0]$ and renormalize them to have mean 1.0 for numerical
stability. This weighting is applied uniformly across all architectures.

We evaluate all models using the Mean Absolute Percentage Error (MAPE) computed in linear SNR space on the
held-out test set as the main metric
\begin{equation}
    \text{MAPE} = 100 \times \frac{1}{N} \sum_{i=1}^{N}
    \left| \frac{\rho_{\rm pred}^{(i)} - \rho_{\rm true}^{(i)}}{\rho_{\rm true}^{(i)}} \right| \, .
\end{equation}
We also define  
\begin{equation}
    \text{Accuracy} = 100-\mathrm{MAPE},
\end{equation}
(in percentage points) as a complementary metric only, not as a classification-style accuracy metric. Because MAPE carries $\rho_{\rm true}$ in the denominator, it is most sensitive to the faintest sources. We therefore additionally report accuracy conditional on the detection-relevant regime $\rho_{\rm true}>8$ in Section~\ref{sec:best-model-diagnostics}.


Although \textsc{$\Phi$asmaNet} operates as a unified system, the $\Phi$-Refractor and
$\Phi$-Feedback are implemented as independent, switchable components to enable controlled
ablations. We report SNR regression with $\Phi$-Refractor \textbf{ON} as the default
configuration. The corresponding ON/OFF ablation across dataset scales is presented in
Section~\ref{sec:refractor-ablation} (Table~\ref{tab:refractor-ablation}), where the two
settings are found to differ only marginally on the SNR task.

\section{Architectures Comparison Across Dataset Sizes}
\label{sec:architecture_comparison}

To evaluate the relative performance and scalability of the four architectures (MLP, ResNet,
Transformer, and \textsc{$\Phi$asmaNet}), we conducted experiments using datasets of varying sizes: 100k,
500k, and 1M samples. All models were trained under the matched protocol described in
Section~\ref{sec:training}, ensuring a controlled and reproducible comparison. This setup allows us
to investigate how each architecture leverages increasing amounts of training data under the
fixed-configuration surrogate setting studied here.



\begin{table}[htbp]
\centering
\caption{Accuracy (\%) before and after monotonic PCHIP calibration (fit on validation, applied to test), reported side-by-side, for the seed-42 dataset realization and training seed~1337. \textsc{$\Phi$asmaNet} uses the default $\Phi$-Refractor-ON configuration at all sizes. At 100k the OFF variant is marginally better (Table~\ref{tab:refractor-ablation}), so the 100k entry is conservative. }
\label{tab:accuracy-datapoints-prepost}
\begin{tabular}{l
                S[table-format=2.2] S[table-format=2.2]
                S[table-format=2.2] S[table-format=2.2]
                S[table-format=2.2] S[table-format=2.2]}
\toprule
\textbf{Architecture} & \multicolumn{2}{c}{\textbf{100k}} & \multicolumn{2}{c}{\textbf{500k}} & \multicolumn{2}{c}{\textbf{1M}} \\
\cmidrule(lr){2-3} \cmidrule(lr){4-5} \cmidrule(lr){6-7}
 & {\textbf{Pre}} & {\textbf{Post}} & {\textbf{Pre}} & {\textbf{Post}} & {\textbf{Pre}} & {\textbf{Post}} \\
\midrule
\textsc{$\Phi$asmaNet}   & {\bfseries 98.32} & {\bfseries 98.37} & {\bfseries 98.92} & {\bfseries 99.16} & {\bfseries 98.73} & {\bfseries 99.29} \\
Transformer    & 96.39 & 96.79 & 97.04 & 97.43 & 96.03 & 97.40 \\
ResNet         & 97.38 & 97.63 & 97.73 & 97.90 & 97.96 & 98.06 \\
MLP            & 95.62 & 96.14 & 94.67 & 95.98 & 94.49 & 95.64 \\
\bottomrule
\end{tabular}
\end{table}

\begin{figure}[htbp]
    \centering
    \includegraphics[width=0.8\linewidth]{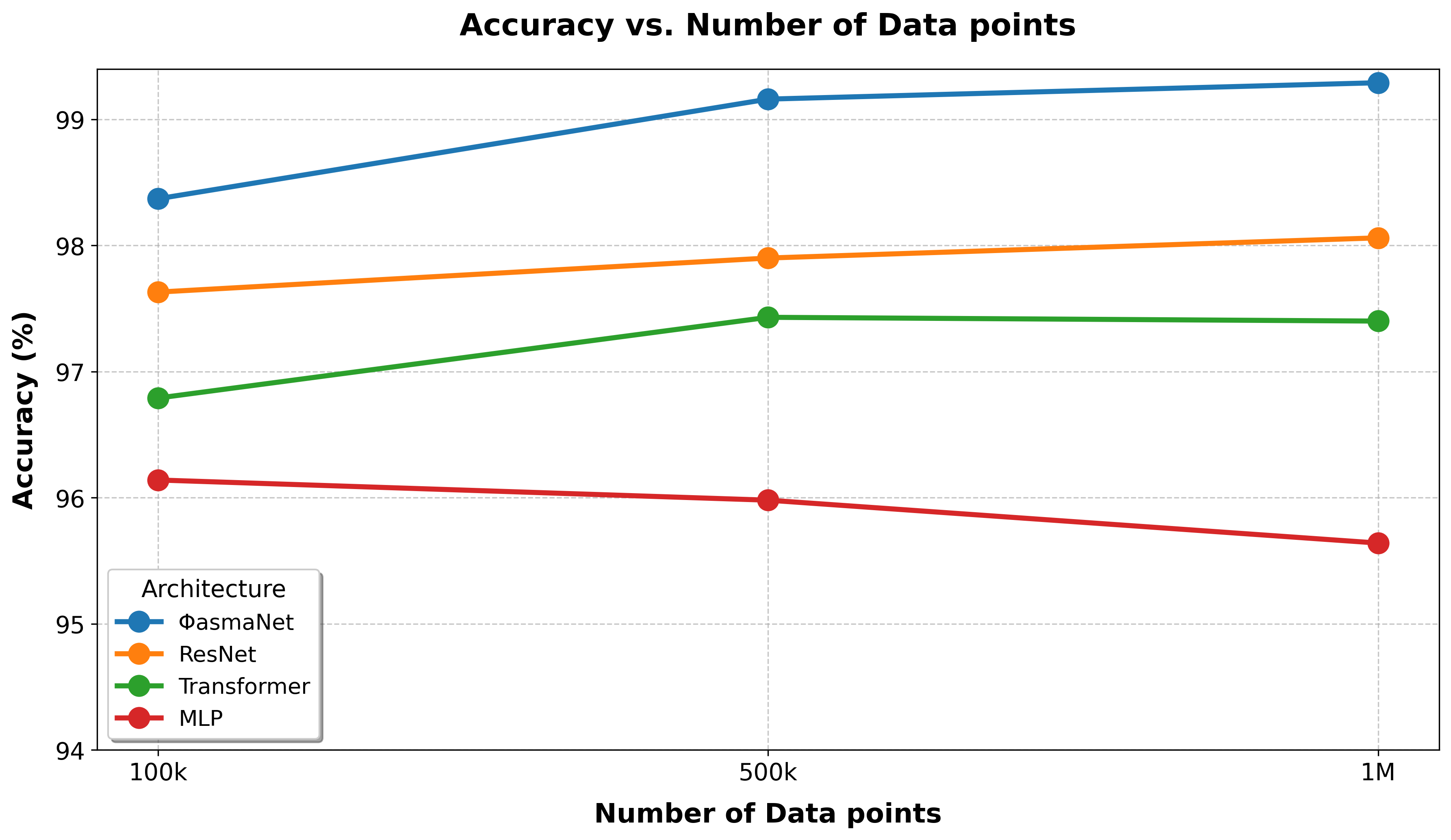}
    \caption{Post-calibration accuracy versus dataset size for all architectures. }
    \label{fig:accuracy-line}
\end{figure}

Table~\ref{tab:accuracy-datapoints-prepost}  summarizes the
test-set accuracy across dataset sizes, reported before and after monotonic
calibration, respectively (for data set realization with seed 42 and training seed 1337). After calibration, \textsc{$\Phi$asmaNet} shows the strongest large-scale
performance, reaching $99.16\%$ at 500k and $99.29\%$ at 1M. ResNet improves to $98.06\%$ at 1M
after calibration, while the Transformer reaches $97.40\%$ and the MLP $95.64\%$ at 1M, see also  Figure~\ref{fig:accuracy-line}. Overall,
\textsc{$\Phi$asmaNet} is the best-performing model across all evaluated dataset sizes under the shared
protocol. We note a slight pre-calibration dip for \textsc{$\Phi$asmaNet} from 500k to 1M
(Table~\ref{tab:accuracy-datapoints-prepost}). The effect is small ($\approx0.2$
percentage points) and confined to the pre-calibration metric: after the rank-preserving monotonic
PCHIP calibration (fit on validation only), which we treat as the primary deployment metric, accuracy
increases monotonically with dataset size (Table~\ref{tab:accuracy-datapoints-prepost}). We do not
over-interpret this small pre-calibration non-monotonicity.

\begin{figure}[htbp]
\centering
\includegraphics[width=0.95\textwidth]{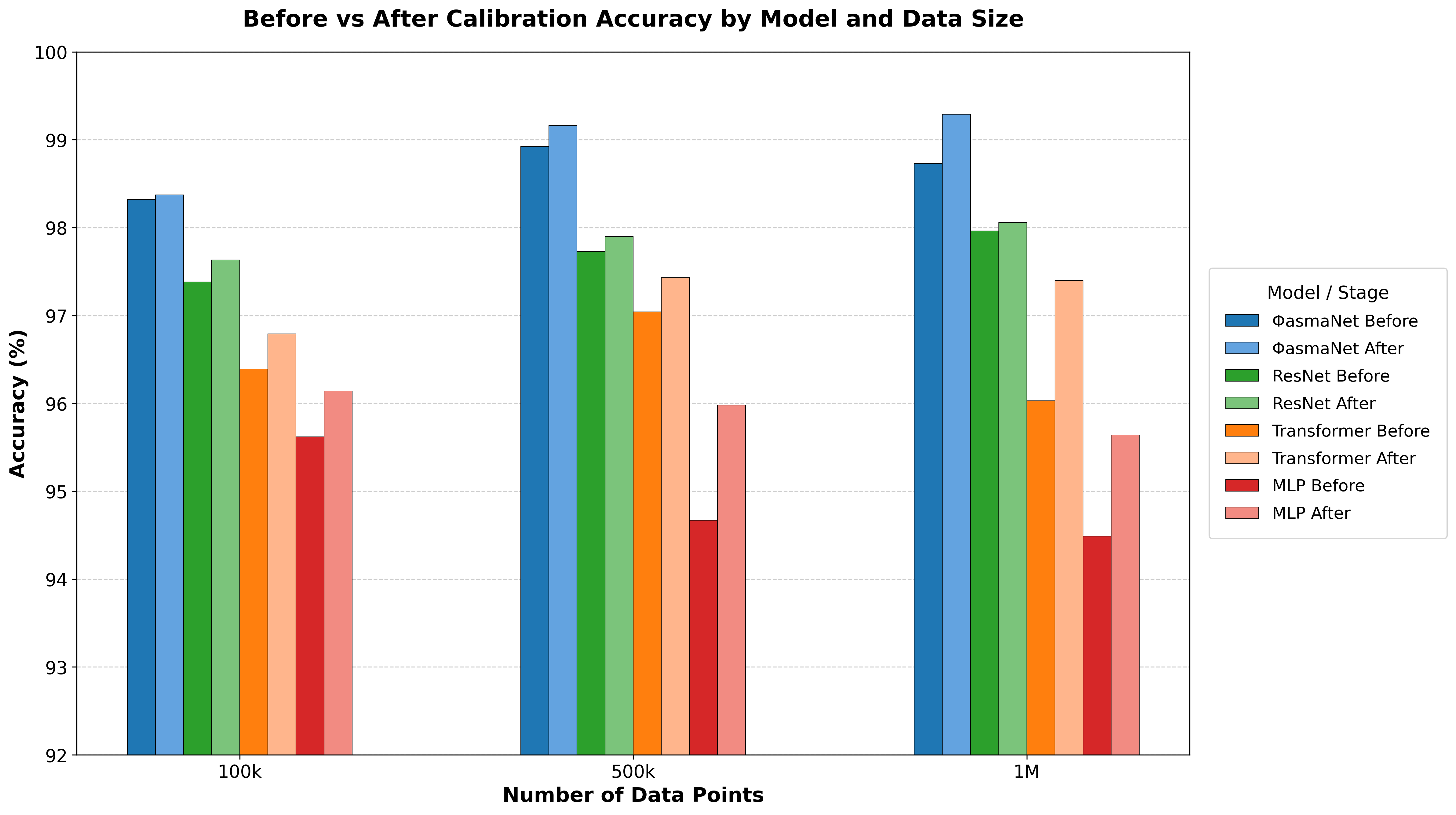}
\caption{Before vs.\ after calibration performance ($100-\mathrm{MAPE}$, \%) by model and dataset
size. The same monotonic PCHIP calibration procedure is fitted on the validation set and applied
uniformly to all models. Differences between pre- and post-calibration bars indicate how much each
model benefits from this lightweight correction under the simulated setting considered here.}
\label{fig:before-after-calibration}
\end{figure}

As shown in Figure~\ref{fig:before-after-calibration}, calibration has a dataset-size-dependent
effect across models. For \textsc{$\Phi$asmaNet}, calibration produces a small change at 100k but a sizeable
improvement at larger scales (500k and 1M), while the other architectures benefit from the same monotonic
correction to varying degrees. 

 As a
lightweight robustness check on the 100k setting, we repeated training with a different internal
training seed (2024) while keeping the dataset generation fixed (seed~42) and the protocol unchanged. Note that
this check uses the $\Phi$-Refractor OFF setting of \textsc{$\Phi$asmaNet}, since at the 100k setting it still performs better than the default ON setting.  The resulting test-set accuracy (pre-/post-calibration) was: \textsc{$\Phi$asmaNet} 98.39/98.49, Transformer
96.41/96.64, ResNet 97.58/97.70, and MLP 95.73/96.34, see Table \ref{tab:robustness-checks}, confirming that the performance ordering of the different architectures is
stable under this seed change.
A dataset-realization check at 1M for \textsc{$\Phi$asmaNet} using additional generation seeds 43 and 44 yielded accuracies of
$98.72\%/99.17\%$ and $98.70\%/99.29\%$ (pre-/post-calibration), respectively, under the same
protocol.

These results suggest that \textsc{$\Phi$asmaNet} is the best-performing model \emph{among the
evaluated architectures} in the 100k--1M regime under the matched protocol considered here.

\begin{table}[htbp]
\centering
\caption{Robustness checks on test accuracy ($100-\mathrm{MAPE}$, \%, pre-/post-calibration): a training-seed change at 100k (all architectures, with \textsc{$\Phi$asmaNet} using $\Phi$-Refractor OFF, the better setting at that scale) and two additional dataset realizations at 1M (generation seeds~43 and~44, \textsc{$\Phi$asmaNet}).}
\label{tab:robustness-checks}
\begin{tabular}{l l S[table-format=2.2] S[table-format=2.2]}
\toprule
\textbf{Setting} & \textbf{Architecture} & {\textbf{Pre}} & {\textbf{Post}} \\
\midrule
\multicolumn{4}{l}{\emph{100k (seed-change check)}} \\
\addlinespace[0.25em]
Train seed~2024 & \textsc{$\Phi$asmaNet} (Refractor OFF) & 98.39 & 98.49 \\
Train seed~2024 & Transformer & 96.41 & 96.64 \\
Train seed~2024 & ResNet & 97.58 & 97.70 \\
Train seed~2024 & MLP & 95.73 & 96.34 \\
\addlinespace[0.35em]
\multicolumn{4}{l}{\emph{1M (dataset-realization check)}} \\
\addlinespace[0.25em]
Gen seed~43 & \textsc{$\Phi$asmaNet} & 98.72 & 99.17 \\
Gen seed~44 & \textsc{$\Phi$asmaNet} & 98.70 & 99.29 \\
\bottomrule
\end{tabular}
\end{table}

\subsection{$\Phi$-Refractor ablation}
\label{sec:refractor-ablation}

Table~\ref{tab:refractor-ablation} isolates the contribution of the $\Phi$-Refractor on the
SNR task by comparing the default ON configuration against an OFF variant under an otherwise
identical protocol (same dataset realization, splits, preprocessing, loss, weighting,
optimizer, schedule, early stopping, and post-hoc calibration). Before calibration the
ablation shows a regime-dependent trend: $\Phi$-Refractor OFF is better at 100k (by
$0.24$~pp), whereas ON is better at 500k and 1M (by $0.19$ and $0.12$~pp). The monotonic PCHIP
calibration we adopt as the primary deployment metric compresses these gaps at the larger
scales: post-calibration, OFF remains ahead at 100k ($+0.20$~pp) and marginally at 500k
($+0.06$~pp), while ON leads only at 1M ($+0.06$~pp). For context, the seed-to-seed variation
of a single configuration is of the same order (the OFF@100k accuracy shifts by
${\sim}0.1$--$0.2$~pp between training seeds~1337 and~2024, Table~\ref{tab:robustness-checks}).
We therefore regard the 500k and 1M differences as within run-to-run variation, and the only
robust SNR-task signal as a modest OFF advantage at 100k. We nonetheless adopt $\Phi$-Refractor
ON as the default \textsc{$\Phi$asmaNet} configuration, motivated by its consistent and larger
benefit on the public tabular benchmarks (e.g.\ Covertype, $0.9634\!\to\!0.9664$,
Table~\ref{tab:covertype-results}), where the effect clearly exceeds the SNR-task noise level.

\begin{table}[htbp]
\centering
\caption{$\Phi$-Refractor ablation (ON vs.\ OFF) on the SNR task: test accuracy
($100-\mathrm{MAPE}$, \%) before and after monotonic PCHIP calibration, under the otherwise
matched protocol of Section~\ref{sec:training}. ON is the default \textsc{$\Phi$asmaNet}
configuration. All entries use generation seed~42 and training/split seed~1337.}
\label{tab:refractor-ablation}
\begin{tabular}{l
                S[table-format=2.2] S[table-format=2.2]
                S[table-format=2.2] S[table-format=2.2]
                S[table-format=2.2] S[table-format=2.2]}
\toprule
\textbf{Configuration} & \multicolumn{2}{c}{\textbf{100k}}
                       & \multicolumn{2}{c}{\textbf{500k}}
                       & \multicolumn{2}{c}{\textbf{1M}} \\
\cmidrule(lr){2-3} \cmidrule(lr){4-5} \cmidrule(lr){6-7}
 & {\textbf{Pre}} & {\textbf{Post}} & {\textbf{Pre}} & {\textbf{Post}} & {\textbf{Pre}} & {\textbf{Post}} \\
\midrule
$\Phi$-Refractor ON  & 98.32 & 98.37 & 98.92 & 99.16 & 98.73 & 99.29 \\
$\Phi$-Refractor OFF & 98.56 & 98.57 & 98.73 & 99.22 & 98.61 & 99.23 \\
\bottomrule
\end{tabular}
\end{table}

\section{Best-model Diagnostics}
\label{sec:best-model-diagnostics}

Having established \textsc{$\Phi$asmaNet} as the best-performing architecture under the matched
protocol in the 100k--1M regime (Section~\ref{sec:architecture_comparison}), we report a compact set
of diagnostics for the best model on the 1M setting (dataset seed~42, split/training seed~1337).
These plots complement the aggregate metrics by characterizing training stability, residual structure
in linear-SNR space, and the effect of the monotonic calibration used throughout this paper.

\subsection{Training dynamics and convergence}
\begin{figure}[htbp]
    \centering
    \includegraphics[width=0.80\linewidth]{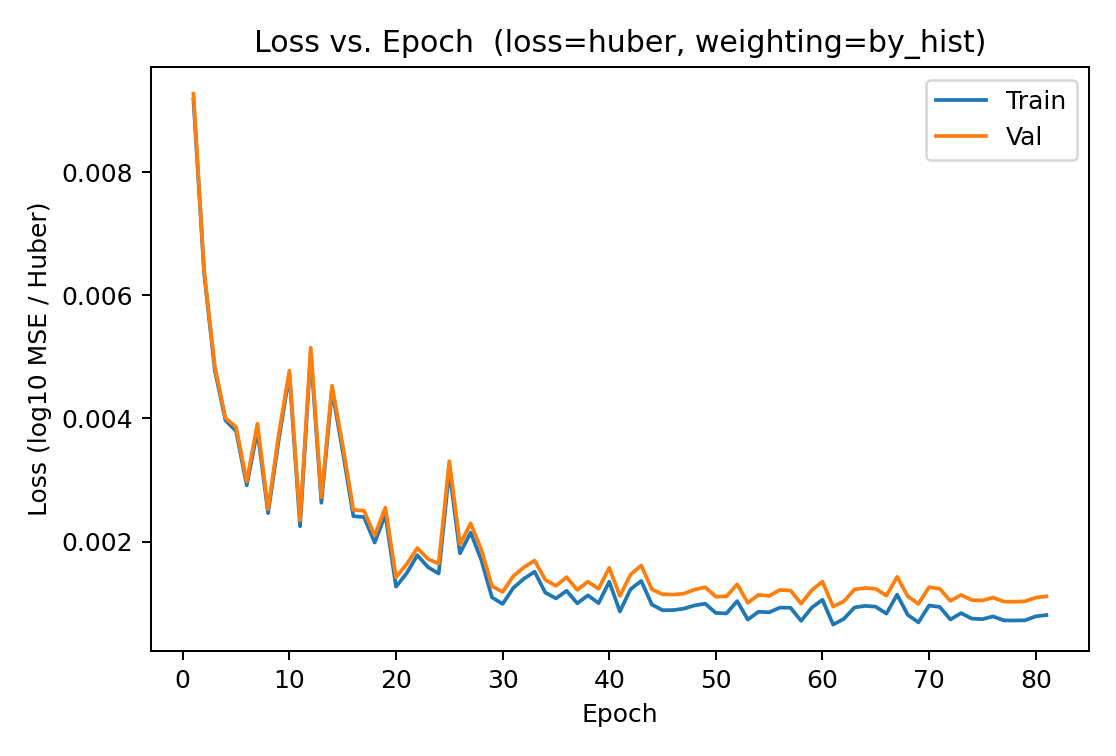}
    \caption{Training and validation loss versus epoch for \textsc{$\Phi$asmaNet} on the 1M dataset
    (seed~42, split/training seed~1337), using Huber loss in $\log_{10}\rho$ space with
    histogram-based weighting \texttt{by\_hist}. The trajectories show rapid early improvement
    followed by a stable plateau under the shared learning-rate scheduling and early-stopping
    protocol.}
    \label{fig:cf-loss-vs-epoch-1m}
\end{figure}

Figure~\ref{fig:cf-loss-vs-epoch-1m} reports the evolution of the training and validation loss
across epochs. The loss decreases rapidly in early epochs and then approaches a stable plateau,
indicating effective optimization under the shared training configuration (AdamW,
ReduceLROnPlateau, and early stopping). The train--validation gap remains small and stable,
consistent with controlled generalization under the fixed regularization settings.

Training and validation losses closely track each
other during the first epochs. This behavior is expected in the 1M-sample regime under an
independent and identically distributed random split, where the model is initially underfitting and the error is bias-dominated rather than
variance-dominated. Moreover, the validation set is large, providing a low-variance estimate of the
generalization loss. Regularization (dropout and weight decay), together with learning-rate
scheduling and early stopping, further limits the train--validation gap. {\it We note that the training
loss is evaluated using the same histogram-based sample weights used for optimization (normalized to
unit mean), while validation loss is reported unweighted}. Their close agreement indicates stable
optimization without evidence of pathological overfitting.

\begin{figure}[htbp]
    \centering
    \includegraphics[width=0.80\linewidth]{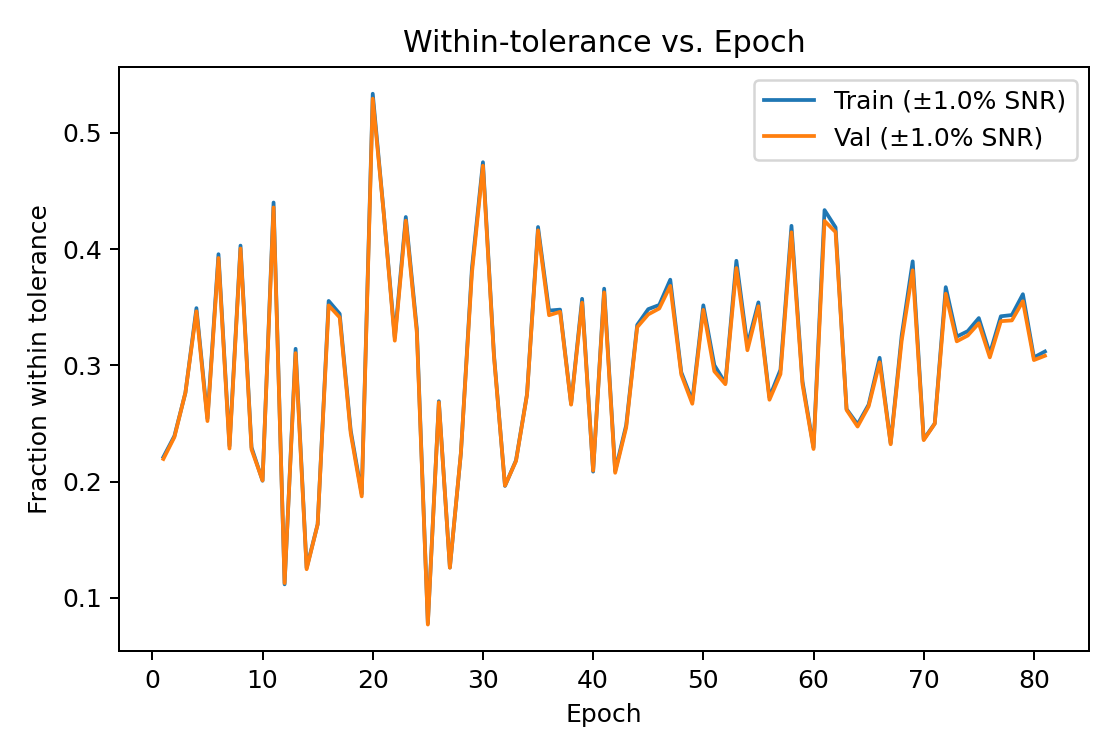}
    \caption{Within-tolerance fraction versus epoch for \textsc{$\Phi$asmaNet} on the 1M dataset
    (seed~42, split/training seed~1337). We report the fraction of samples with
    $\left|\rho_{\mathrm{pred}}-\rho_{\mathrm{true}}\right|/\rho_{\mathrm{true}} \le 1\%$ in
    linear-SNR space on train and validation splits. This strict metric stabilizes at later epochs,
    consistent with convergence under the matched training protocol.}
    \label{fig:cf-within-tol-vs-epoch-1m}
\end{figure}

Figure~\ref{fig:cf-within-tol-vs-epoch-1m} complements this view with an application-facing metric:
the fraction of samples within $\pm 1\%$ relative error in linear-SNR space. This strict
within-tolerance curve is more sensitive to tail events and target imbalance (amplified by
histogram-based weighting), and therefore exhibits larger fluctuations than the averaged loss.
Nevertheless, it stabilizes in later epochs, consistent with convergence of the fitted surrogate.

\subsection{Residual structure and the effect of calibration}
\begin{figure}[htbp]
    \centering
    \includegraphics[width=0.49\linewidth]{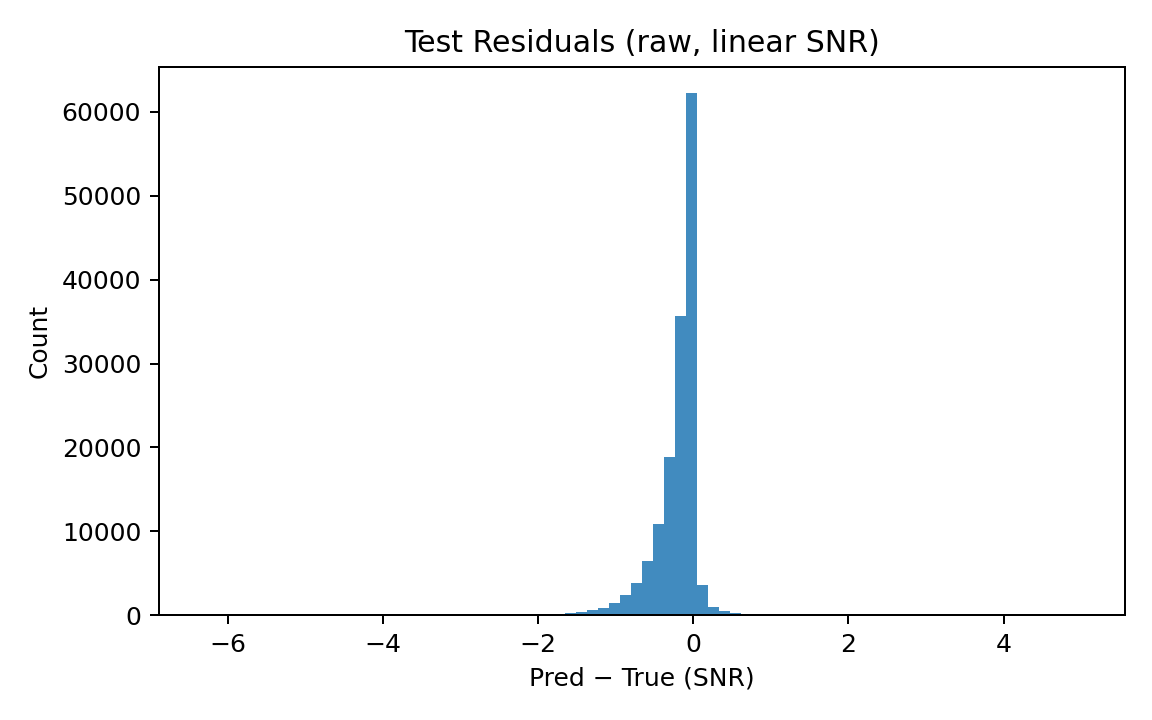}
    \hfill
    \includegraphics[width=0.49\linewidth]{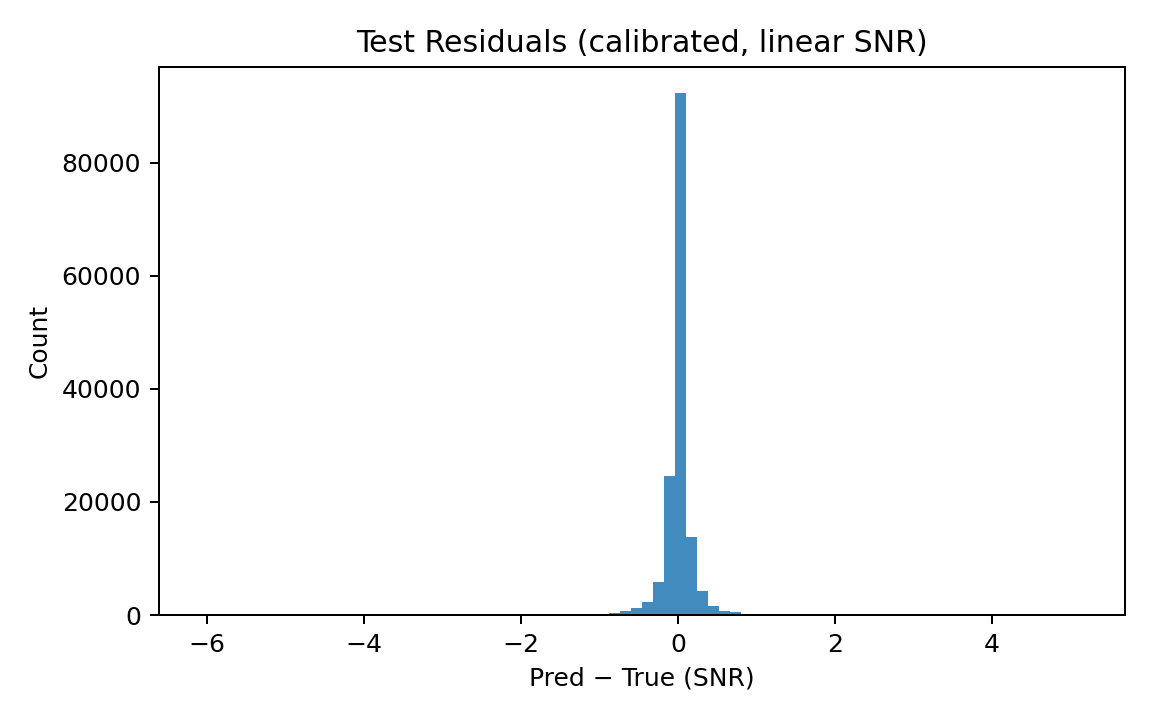}
    \caption{Test residual histograms for \textsc{$\Phi$asmaNet} on the 1M dataset (seed~42,
    split/training seed~1337) in linear-SNR space. \textbf{Left:} raw residuals
    ($\rho_{\mathrm{pred}}-\rho_{\mathrm{true}}$). \textbf{Right:} residuals after monotonic PCHIP
    calibration (fit on validation, applied to test).}
    \label{fig:cf-residual-hist-raw-cal-1m}
\end{figure}

To directly visualize the effect of the monotonic PCHIP calibration (fit on validation, applied to
test), Figure~\ref{fig:cf-residual-hist-raw-cal-1m} compares test residual histograms in linear-SNR
space before and after calibration. The calibrated residuals are more tightly concentrated around
zero, indicating a reduction in global systematic bias while preserving rank ordering by construction.

\begin{figure}[htbp]
    \centering
    \includegraphics[width=0.80\linewidth]{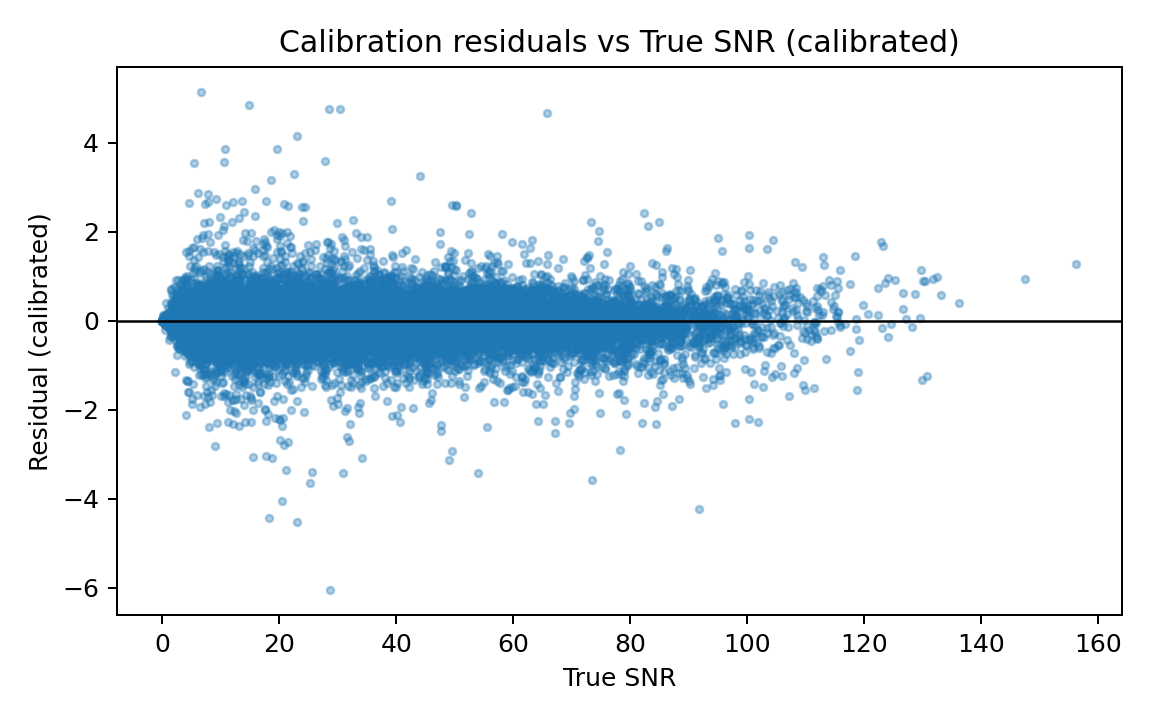}
    \caption{Calibrated test residuals versus true SNR for \textsc{$\Phi$asmaNet} on the 1M dataset
    (seed~42, split/training seed~1337), shown in linear-SNR space. This plot checks for
    SNR-dependent structure after applying the monotonic PCHIP mapping (fit on validation, applied
    to test).}
    \label{fig:cf-residual-vs-true-1m}
\end{figure}

Beyond marginal residual distributions, it is important to check for SNR-dependent structure that
could matter for selection-effect workflows. Figure~\ref{fig:cf-residual-vs-true-1m} plots calibrated
test residuals versus the true SNR in linear space. Residuals remain centered close to zero across
the bulk of the SNR range, suggesting that calibration removes most global bias without inducing a
strong SNR-dependent drift. The remaining spread reflects heteroscedasticity expected in this
regression setting.

\begin{figure}[htbp]
    \centering
    \includegraphics[width=0.80\linewidth]{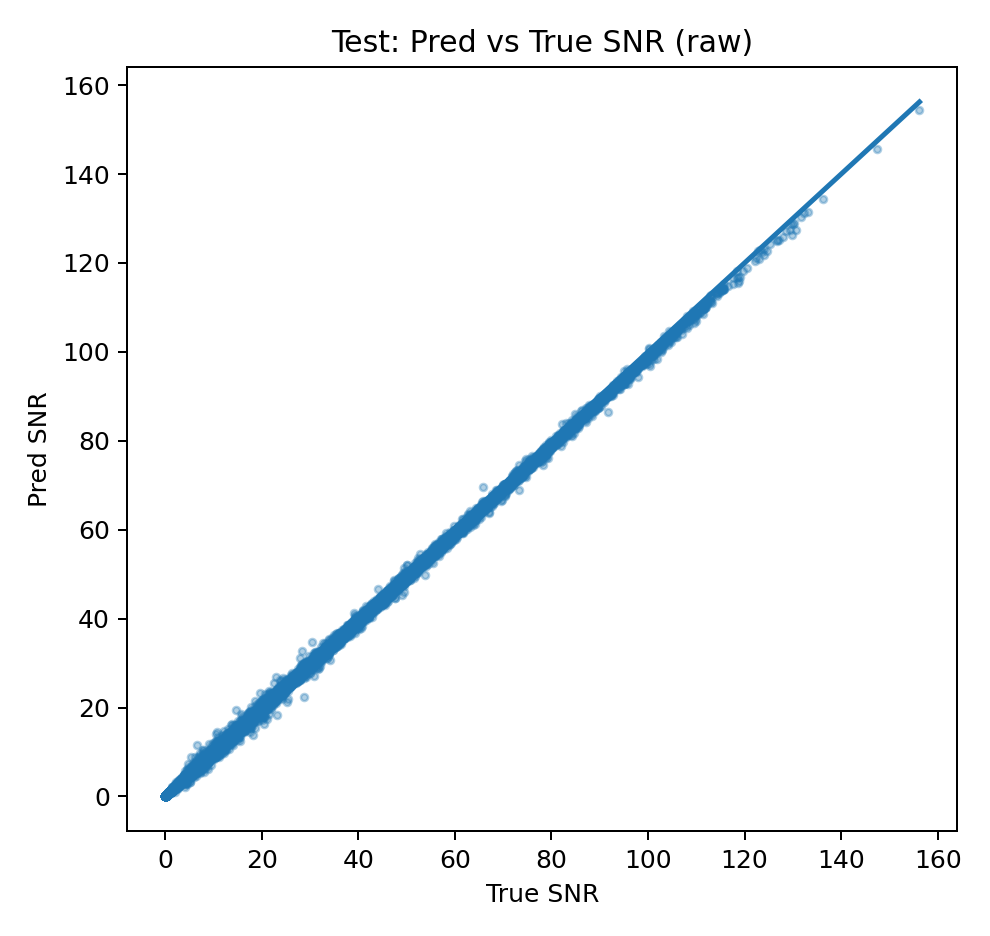}
    \caption{Predicted versus true SNR for \textsc{$\Phi$asmaNet} on the 1M test set (seed~42,
    split/training seed~1337), shown in linear-SNR space (raw predictions).}
    \label{fig:cf-scatter-raw-1m}
\end{figure}

\begin{figure}[htbp]
    \centering
    \includegraphics[width=0.80\linewidth]{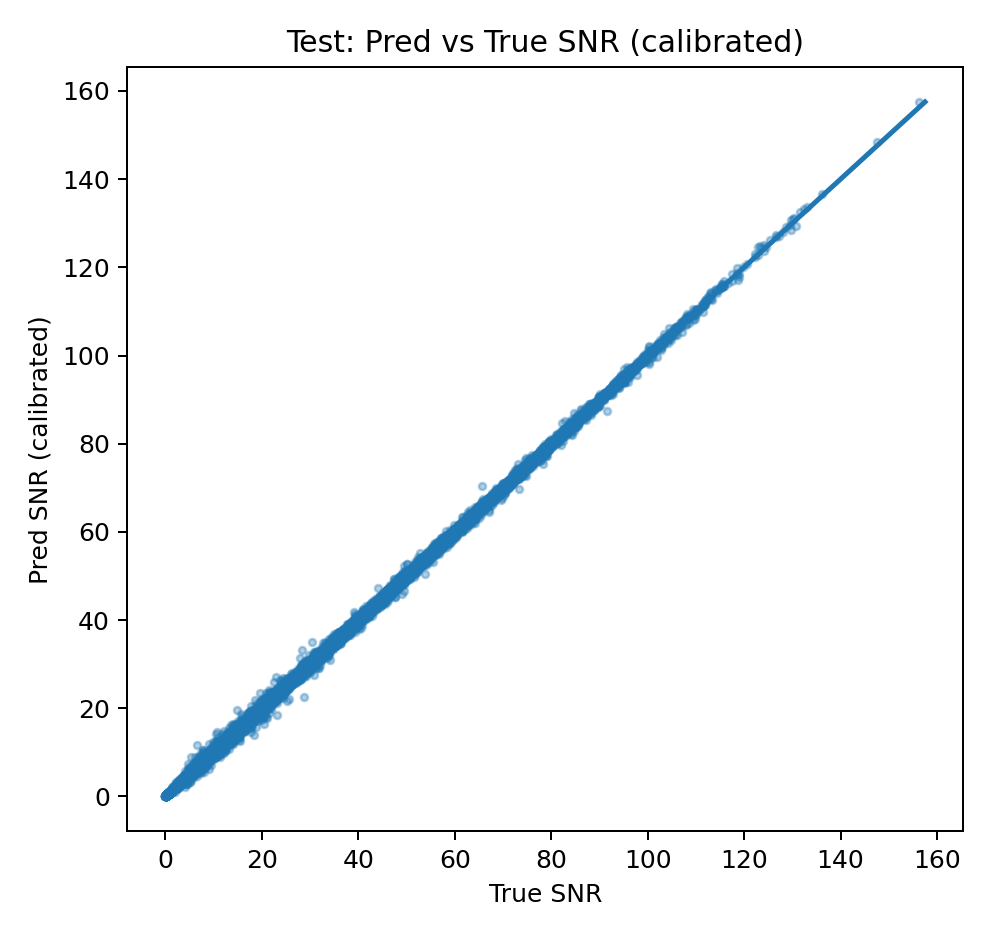}
    \caption{Predicted versus true SNR for \textsc{$\Phi$asmaNet} on the 1M test set (seed~42,
    split/training seed~1337), shown in linear-SNR space after monotonic PCHIP calibration (fit on
    validation, applied to test).}
    \label{fig:cf-scatter-cal-1m}
\end{figure}

Finally, Figures~\ref{fig:cf-scatter-raw-1m}--\ref{fig:cf-scatter-cal-1m} provide a qualitative
check via predicted-versus-true scatter plots in linear-SNR space, before and after calibration. The
calibrated mapping preserves monotone ordering while reducing systematic deviations, consistent with
the post-calibration improvements reported in Section~\ref{sec:architecture_comparison}.

\subsection{Accuracy in the detection-relevant regime}
Because MAPE is a relative-error metric, samples with very small
$\rho_{\rm true}$ can numerically dominate the aggregate error. Following
common practice in selection-effect analyses~\cite{mould2024calibrating}, we additionally
report accuracy conditional on the canonical single-detector detection
threshold $\rho_{\rm true} > 8$. On the 1M test set (dataset seed~42,
split/training seed~1337), this subset comprises $N = 96{,}047$ samples
($64.0\%$ of the held-out test set). Under the same monotonic PCHIP
calibration procedure described in Section~\ref{sec:training} (fit on the
validation split, applied to the test split), \textsc{$\Phi$asmaNet} attains
$100{-}\mathrm{MAPE} = 98.71\%$ pre-calibration and $\mathbf{99.39\%}$
post-calibration on this subset, with $R^2 = 0.9998$ in linear-SNR space.
The detection-relevant accuracy is thus comparable to the full-set value
pre-calibration ($98.71\%$ vs.\ $98.73\%$) and slightly higher post-calibration
($99.39\%$ vs.\ $99.29\%$), confirming that the
surrogate's predictive quality in the astrophysically detectable regime is
at least as high as on the full distribution.

\subsection{Computational efficiency}
To quantify end-to-end inference cost, we perform a full evaluation pass
on an independent $10^6$-sample dataset using a saved checkpoint from the 1M-sample training run. The
evaluation was executed on a single GPU (RTX 4080) with batch size $B=1024$. The total wall-clock
time for the complete pass, including data loading, forward passes, metric computation, and calibration, was
approximately 8.3\,s, corresponding to an effective throughput of ${\approx}\,1.2 \times 10^5$
  samples per second. The corresponding wall-clock time of a traditional calculation of the SNR was 980s on a 16-core CPU with parallel execution. Using \textsc{$\Phi$asmaNet} on a modern GPU is therefore about {\it two orders of
  magnitude} faster than the traditional SNR computation on a 16-core CPU, or {\it more than three orders of magnitude} faster than a single-core CPU implementation. This demonstrates that \textsc{$\Phi$asmaNet} can provide rapid SNR predictions suitable for large-scale simulation and selection-effect studies.



Turning from throughput to model size, Table~\ref{tab:param-counts} summarizes the total number of trainable parameters
for each architecture under the default hyperparameter configurations used
throughout this work (see Sections~\ref{sec:mlp}--\ref{sec:phasmanet}).
\textsc{$\Phi$asmaNet} and the Transformer encoder are the most compact models,
with approximately $732$k and $540$k parameters, respectively.  The MLP baseline
is the lightest at roughly $201$k parameters owing to its simple stacked-linear
design ($11 \!\to\! 256 \!\to\! 256 \!\to\! 256 \!\to\! 256 \!\to\! 1$), while
the 1D-ResNet is the largest at approximately $3.98$M parameters, dominated by
its deepest convolutional stage (Stage~4, $C{=}512$), which alone accounts for
roughly $73\%$ of the total.  Notably, the recursive $\Phi$-Feedback passes in
\textsc{$\Phi$asmaNet} reuse the same $\Phi$-Stage and gate weights at every
iteration, so increasing $R$ adds \emph{no} extra parameters. The computational
cost grows only in FLOPs per forward pass.  Overall,
\textsc{$\Phi$asmaNet} achieves the best accuracy--parameter trade-off among the
evaluated architectures: with ${\sim}\,5.4{\times}$ fewer parameters than the
ResNet and only ${\sim}\,1.4{\times}$ more than the Transformer, it attains the
highest post-calibration $100{-}\mathrm{MAPE}$ at every dataset scale considered
(Section~\ref{sec:architecture_comparison}).

\begin{table}[t]
\centering
\caption{Trainable parameter counts for each architecture under the default
configurations used in this work.  All counts refer to the models as
instantiated for the 11-feature SNR regression task.}
\label{tab:param-counts}
\small
\begin{tabular}{@{}l r r l@{}}
\toprule
\textbf{Architecture} & \textbf{Params} & \textbf{Relative} & \textbf{Key sizing}\\
\midrule
MLP              & $\sim$201\,k  & $1.0\times$   & width $w{=}256$, depth $d{=}4$\\
Transformer      & $\sim$540\,k  & $2.7\times$   & $d_\mathrm{model}{=}128$, $N_L{=}4$, $N_h{=}8$, $d_\mathrm{ff}{=}256$\\
\textsc{$\Phi$asmaNet}    & $\sim$732\,k  & $3.7\times$   & $T{=}8$, $D{=}128$, $K{=}3$, $R{=}2$, $P{=}4$\\
ResNet (1D)      & $\sim$3.98\,M & $19.8\times$  & width $w{=}64$, 4 stages $\times$ 2 blocks\\
\bottomrule
\end{tabular}
\end{table}







\subsection{From SNR Predictions to Detection Probability}
\label{sec:popdet}
The application that motivates a fast SNR surrogate is the estimation of selection effects
in population studies, where the detectability of every source in a large synthetic
catalogue must be evaluated, and re-evaluated at each step of a hierarchical inference
\cite{gerosa2020gravitational,chapman2023rapid,callister2024}. The single-detector
restriction and fixed reference distance adopted here are not obstacles to this use: because
the optimal SNR scales as $\rho\propto1/d_L$, a label computed at $d_L^{\rm ref}=1000$~Mpc
can be rescaled to any luminosity distance at query time,
$\rho(d_L)=\rho^{\rm ref}\,d_L^{\rm ref}/d_L$, so a surrogate trained at a single reference
distance already returns the SNR for a population drawn over distance.

In the present work we adopt the simplest detectability model, a hard threshold on the
optimal SNR,
\begin{equation}
P_{\mathrm{det}}(\bm\theta)=P(\mathrm{det}\mid\bm\theta)
=\Theta\!\left(\rho(\bm\theta)-\rho_{\mathrm{th}}\right),
\qquad \rho_{\mathrm{th}}=8,
\label{eq:pdet}
\end{equation}
not to carry out a population analysis, but as an additional, detection-oriented check on
the surrogate. Applying Eq.~(\ref{eq:pdet}) to the predicted and to the true SNRs assigns
each test source a binary detectability label, and the two labels can be compared directly.
Mislabellings can occur only for sources whose true SNR falls within the surrogate's error
band of the threshold, so the strong near-threshold fidelity reported above ($R^2=0.9998$
and $100-\mathrm{MAPE}=99.39\%$ on the $\rho_{\rm true}>8$ subset) implies that the predicted
and reference detection labels coincide for all but a small fraction of sources near
$\rho_{\rm th}$, and hence that the predicted detectable fraction closely tracks the
ground-truth value ($\approx64\%$ at the reference distance).

We stress that Eq.~(\ref{eq:pdet}) is a deterministic, optimal-SNR proxy, used here purely
as an accuracy diagnostic. A realistic detection probability instead averages over noise
realizations, modelling the observed SNR as a non-central $\chi^2$ variate about $\rho$, and, ideally, over search-pipeline effects such as non-stationarity and
signal-consistency cuts that a single SNR threshold cannot capture
\cite{callister2024,mould2024calibrating,gwsnr}. Embedding \textsc{$\Phi$asmaNet} in such a
noise-averaged selection-function pipeline, and quantifying the resulting bias on the
sensitive volume $V\!T$ and on inferred population hyperparameters, is the subject of
planned follow-up work. Here we report Eq.~(\ref{eq:pdet}) only as a complementary check on
the surrogate's accuracy in the detection-relevant regime.

\section{Discussion}
\label{sec:discussion}

We introduced \textsc{$\Phi$asmaNet}, a staged spectral token-refinement architecture with $\Phi$-Stages, the
$\Phi$-Refractor subspace mixing module, and $\Phi$-Feedback top-down gated refinement, evaluated as
a surrogate model for fast SNR prediction in a controlled, simulation-based setting. The learning
task is to map a fixed-dimensional vector of intrinsic and extrinsic parameters (including distance
and antenna-pattern projections) to the phase-averaged optimal PyCBC SNR under a fixed representative
PSD and waveform approximant. We do not infer SNR from noisy strain time series.

Across dataset scales from 100k to 1M simulated binary black hole injections,
\textsc{$\Phi$asmaNet} consistently outperforms the evaluated Transformer-, ResNet-, and MLP-based
\emph{neural} baselines under the matched protocol described in Section~\ref{sec:training}. At the 1M-sample scale, \textsc{$\Phi$asmaNet} reaches $100-\mathrm{MAPE}=98.73\%$
before calibration and $99.29\%$ after monotonic PCHIP calibration on the held-out test set under
the simulated configuration considered here. These figures are stable under
dataset-realization variation: two additional independently generated 1M datasets (generation
seeds 43 and 44) yield $98.72\%/99.17\%$ and $98.70\%/99.29\%$ pre-/post-calibration
(Section~\ref{sec:architecture_comparison}). They are obtained with a compact model of
${\sim}\,732$k trainable parameters, the most favorable accuracy--parameter trade-off among the
evaluated architectures (Table~\ref{tab:param-counts}).

With a sustained throughput of ${\sim}\,1.2\times10^{5}$ SNR evaluations per second on a single GPU (a $10^{6}$-injection set processed end-to-end in ${\sim}\,8.3$~s, Section~\ref{sec:best-model-diagnostics}), orders of magnitude faster than traditional methods, the approach is suitable
for high-throughput simulation workflows under the assumptions described in
Section~\ref{sec:methodology}. Accuracy is highest precisely where these
workflows need it: conditional on the detection-relevant regime $\rho_{\rm true}>8$, the calibrated
surrogate attains $100-\mathrm{MAPE}=99.39\%$ with $R^2=0.9998$ in linear-SNR space
(Section~\ref{sec:best-model-diagnostics}).

These results support the use of \textsc{$\Phi$asmaNet} as a compact surrogate for fast SNR evaluation in
simulation-driven workflows where many SNR computations are required. The primary intended application
is the estimation of selection effects in population inference, where the detectability of every
source in a large synthetic catalogue must be evaluated repeatedly. The $\rho\propto1/d_L$ rescaling
and threshold-based $P_{\rm det}$ construction of Section~\ref{sec:popdet} make the present
single-distance surrogate directly usable for this purpose.

A direct, apples-to-apples comparison against earlier neural SNR surrogates is not currently possible:
prior studies adopt smaller input parametrizations (with correspondingly stronger approximations),
different parameter ranges, and different detector noise curves, so no common external benchmark
exists. Rather than claim superiority over such heterogeneous setups, we therefore report performance
only against neural baselines trained under an identical, fully specified protocol. This
study documents the full surrogate pipeline (sampling priors, waveform settings, data splits,
training protocol, and calibration) at a level intended to make it reproducible end-to-end.
Together with our public code release (Section~\ref{sec:code}), it is intended to provide a
concrete, well-defined benchmark against which future surrogates, including non-neural tabular
regressors such as gradient-boosted trees and packages such as \textsc{gwsnr} \cite{gwsnr}, can be
compared.

We also stress the scope of the present study. The surrogate is trained for a single detector
(LIGO-Hanford) under a fixed representative PSD and waveform approximant, operates on a
$11$-dimensional parameter/projection vector rather than detector strain, and is fitted over a bounded
operating range ($\rho<200$). Within these assumptions it is accurate and fast, but its predictions
are tied to the configuration used to generate the labels: transfer to other PSDs, observing epochs,
or detector networks is not established here, and end-to-end detection from noisy strain is a
different problem that a parameter-to-SNR surrogate does not address. Future work will therefore
concentrate on extensions that remain within the surrogate setting while broadening its reach.
First, we will cover detector networks by training one surrogate per detector and combining the per-detector
SNRs in quadrature (Section~\ref{sec:methodology}), together with retraining for, or conditioning
on, the PSDs of different observing epochs. Second, we will replace the hard $\rho_{\rm th}=8$ proxy with a
noise-averaged detection probability and quantify the induced bias on the sensitive volume $V\!T$
and on inferred population hyperparameters, as outlined in Section~\ref{sec:popdet}.
Third, we will broaden the architectural ablations and add strong non-neural tabular baselines, such as
gradient-boosted trees, to isolate which components drive the observed gains. Finally, we will assess
\textsc{$\Phi$asmaNet} as a general-purpose architecture for structured tabular prediction, for which the
out-of-the-box benchmarks of Appendix~\ref{sec:appendix} provide encouraging preliminary evidence.

\section{Code Availability}
\label{sec:code}

The training code for \textsc{$\Phi$asmaNet} (with and without the $\Phi$-Refractor) and for the
MLP, ResNet and Transformer baselines is publicly available at
\url{https://github.com/giannis00/PhasmaNet-SNR}.

\section{Acknowledgments}

We are grateful to Grigorios Papigkiotis, Alexandra Koloniari and Georgios Vardakas for comments. 
Virgo is funded through the European Gravitational Observatory (EGO), by the French Centre National de Recherche Scientifique (CNRS), the Italian Istituto Nazionale di Fisica Nucleare (INFN) and the Dutch Nikhef, with contributions by institutions from Belgium, Germany, Greece, Hungary, Ireland, Japan, Monaco, Poland, Portugal, Spain. Results presented in this work have been produced using the AUTh Computing Infrastructure and Resources. 

The authors used ChatGPT-5/5.1 and Claude Opus 4.5 in parts of this manuscript for code generation, text generation, enhancing the clarity and structure of the work, and creation of explanatory figures. The authors have critically revised any generated text to ensure it is accurate and free from plagiarism. The authors are responsible for the interpretation, reasoning and conclusions presented in the manuscript. The authors ensured that AI-assisted explanatory figures did not alter, distort or misrepresent the underlying data.

\section{Appendix}
\label{sec:appendix}

The main text evaluates \textsc{$\Phi$asmaNet} as a surrogate for a single, tightly
controlled task: phase-averaged optimal SNR regression under a fixed PSD,
waveform approximant, and detector configuration. A natural concern is
whether the gains we attribute to the staged inductive bias of \textsc{$\Phi$asmaNet}
are specific to this $11$-dimensional, smooth-target regime, or whether they
reflect a more general property of the architecture. To probe this, we
report out-of-the-box results on three standard public tabular benchmarks
spanning regression (YearPredictionMSD), multiclass classification
(Covertype), and binary classification on a high-dimensional dense input
(Epsilon, $2{,}000$ features), and compare against the nine deep tabular
architectures benchmarked under identical splits in~\cite{gorishniy2021revisiting}.

We emphasize that these experiments are deliberately conducted \emph{without}
any dataset-specific hyperparameter tuning: the model configuration, tokenizer,
and training protocol are inherited unchanged from the SNR regression setup
of Section~\ref{sec:phasmanet}, whereas the reference architectures
in~\cite{gorishniy2021revisiting} are individually tuned per-dataset via
Bayesian optimization. The Epsilon benchmark, whose input dimensionality
is two orders of magnitude larger than the regime the \textsc{$\Phi$asmaNet} tokenizer
was sized for, additionally serves as a deliberate stress test of
transferability rather than a like-for-like comparison. Our goal in this
appendix is therefore not to claim state-of-the-art results on standard
tabular benchmarks, but to provide controlled evidence that the
\textsc{$\Phi$asmaNet} inductive bias (staged $\Phi$-Stages, $\Phi$-Refractor
subspace mixing, and $\Phi$-Feedback top-down refinement) transfers
beyond the gravitational-wave SNR surrogate task.

\paragraph{Out-of-the-box performance on YearPredictionMSD (regression).}
To contextualize these numbers, we compare against the nine deep tabular
architectures benchmarked on YearPredictionMSD in \cite{gorishniy2021revisiting}
(MLP, SNN, TabNet, GrowNet, DCN2, AutoInt, NODE, ResNet, and FT-Transformer),
evaluated on the same train/validation/test split we use here. Under our
out-of-the-box configuration, \textsc{$\Phi$asmaNet} (Refractor \textbf{ON}) attains
$\mathrm{RMSE} = 8.8738$, which outperforms \textbf{4 out of 9} reference
architectures (TabNet at $8.909$, SNN at $8.895$, DCN2 at $8.890$, and AutoInt at
$8.882$), placing it at rank \textbf{6/10} (the nine references plus \textsc{$\Phi$asmaNet}) on this benchmark, within
$0.09$ RMSE points of the top deep baseline (NODE at
$8.784$).
We emphasize that this result is obtained \emph{without any hyperparameter tuning}
on YearPredictionMSD, whereas the reference architectures in
\cite{gorishniy2021revisiting} are individually tuned per-dataset via Bayesian
optimization. A task-specific hyperparameter search for \textsc{$\Phi$asmaNet}
would likely improve its relative position further.

\paragraph{Out-of-the-box performance on Covertype (multiclass classification).}
We compare against the deep tabular architectures benchmarked on Covertype in
\cite{gorishniy2021revisiting} under the same split and evaluation protocol. Of
these, eight report a Covertype number (GrowNet does not). Under our out-of-the-box configuration, \textsc{$\Phi$asmaNet} (Refractor
\textbf{ON}) attains a test accuracy of $0.9664$, which outperforms
\textbf{7 out of 8} reference architectures (DCN2 at $0.965$, ResNet at
$0.964$, MLP at $0.962$, SNN at $0.961$, NODE at $0.958$, TabNet at $0.957$,
and AutoInt at $0.934$), placing it at rank \textbf{2/9}. The only
architecture that surpasses it is FT-Transformer ($0.970$).
As in the regression case, this result is obtained without any dataset-specific
hyperparameter search, while the reference architectures in
\cite{gorishniy2021revisiting} are individually tuned per dataset. This reinforces
our view that the observed performance reflects the inductive bias of
\textsc{$\Phi$asmaNet} itself rather than careful tuning.

\begin{table}[t]
\centering
\caption{Covertype test performance of \textsc{$\Phi$asmaNet} (mean $\pm$ SEM over
seeds). No hyperparameter tuning was performed. The configuration is inherited from
the SNR regression experiments in Section~\ref{sec:phasmanet}.}
\label{tab:covertype-results}
\begin{tabular}{lcc}
\toprule
\textbf{\textsc{$\Phi$asmaNet} variant} & \textbf{Accuracy} & \textbf{Cross-entropy} \\
\midrule
$\Phi$-Refractor \textbf{ON}  & $0.9664 \pm 0.0002$ & $0.2828 \pm 0.0049$ \\
$\Phi$-Refractor \textbf{OFF} & $0.9634$            & --                   \\
\bottomrule
\end{tabular}
\end{table}


\paragraph{Out-of-the-box performance on Epsilon (binary classification).}
We evaluate \textsc{$\Phi$asmaNet} on the Epsilon binary classification
benchmark and contextualize the results against the nine deep tabular
architectures reported in \cite{gorishniy2021revisiting}, evaluated under the
same train/validation/test split. Table~\ref{tab:epsilon-results} summarizes
test performance (mean $\pm$ SEM over seeds): \textsc{$\Phi$asmaNet} attains a
scaled log loss of $25.9741 \pm 0.1289$, accuracy $0.8915 \pm 0.0005$, and
$\mathrm{ROC\text{-}AUC} = 0.9950 \pm 0.0003$. The near-perfect ROC-AUC
indicates strong and stable discrimination on this high-dimensional dense
tabular input.

In terms of accuracy, \textsc{$\Phi$asmaNet} places near the bottom of the
ranking on this dataset, outperforming only TabNet ($0.8896$) and trailing
the top tuned architecture (FT-Transformer at $0.8982$). However, the spread
between the best and worst deep architectures on Epsilon is small: all nine
reference architectures lie within approximately $0.9$ percentage points of
each other ($0.8896$--$0.8982$), and \textsc{$\Phi$asmaNet}'s accuracy is
within $0.67$ percentage points of the best tuned model.

As noted above, this is a deliberate stress test rather than a like-for-like
comparison: the tokenizer (a single linear projection to $T=8$ tokens) and the full configuration are
inherited unmodified from the $11$-dimensional SNR task and applied untuned to a $2{,}000$-dimensional
input. Even so, \textsc{$\Phi$asmaNet} remains within the same performance band as the tuned deep
tabular baselines. A dedicated tokenizer and hyperparameter search for high-dimensional dense inputs
would be the natural next step to improve its competitiveness in this regime.

\begin{table}[t]
\centering
\caption{Epsilon test performance of \textsc{$\Phi$asmaNet} (mean $\pm$ SEM over
seeds). Log loss is scaled by $\times 100$ as is customary for this dataset. No
hyperparameter tuning was performed. The configuration is inherited from the SNR
regression experiments in Section~\ref{sec:phasmanet}.}
\label{tab:epsilon-results}
\begin{tabular}{lccc}
\toprule
\textbf{\textsc{$\Phi$asmaNet} variant} & \textbf{LogLoss $\times 100$} & \textbf{Accuracy} & \textbf{ROC-AUC} \\
\midrule
Default (Refractor \textbf{ON}) & $25.9741 \pm 0.1289$ & $0.8915 \pm 0.0005$ & $0.9950 \pm 0.0003$ \\
\bottomrule
\end{tabular}
\end{table}

\paragraph{Discussion of the tabular benchmarks.}
Taken together, the three benchmarks (one regression, two classification) indicate that the
\textsc{$\Phi$asmaNet} inductive bias (staged $\Phi$-Stages, $\Phi$-Refractor subspace mixing,
and $\Phi$-Feedback top-down refinement) transfers beyond the gravitational-wave SNR surrogate
task to standard public tabular problems, \emph{without any dataset-specific hyperparameter tuning}.
On YearPredictionMSD it attains competitive RMSE, landing in the middle of the tuned
reference architectures reported in \cite{gorishniy2021revisiting} (rank 6/10). On Covertype it
attains high multiclass accuracy and low cross-entropy, ranking near the top (rank 2/9, surpassed
only by FT-Transformer), with a modest $\Phi$-Refractor ON/OFF difference
(Table~\ref{tab:covertype-results}) consistent with the $\Phi$-Refractor acting as a task-dependent
representational enhancer. The strong ranking suggests that the $\Phi$-Stage and $\Phi$-Feedback
hierarchy can capture non-linear feature interactions at scale.
On Epsilon, whose $2{,}000$-dimensional input is far outside the regime the \textsc{$\Phi$asmaNet}
tokenizer was designed for, it still achieves strong discrimination (near-perfect AUC) and remains
within roughly $0.7$ percentage points of the best tuned deep baseline. We emphasize that our goal
here is not an exhaustive benchmark study but a controlled demonstration of transferability: a full
per-dataset tuning study, together with a tokenizer variant adapted to high-dimensional dense inputs,
is left for future work and would be the natural next step to establish \textsc{$\Phi$asmaNet} as a
competitive general-purpose tabular architecture.

\bibliographystyle{plain}
\bibliography{sample}

\end{document}